\documentclass[preprint,12pt]{elsarticle}

\usepackage{lineno,hyperref}
\usepackage{bm}
\usepackage{amsfonts}
\usepackage{amssymb}
\usepackage{amsmath}
\usepackage{isomath}

\usepackage{adjustbox}

\usepackage{scalerel,stackengine}
\stackMath
\newcommand\reallywidehat[1]{%
\savestack{\tmpbox}{\stretchto{%
  \scaleto{%
    \scalerel*[\widthof{\ensuremath{#1}}]{\kern-.6pt\bigwedge\kern-.6pt}%
    {\rule[-\textheight/2]{1ex}{\textheight}}
  }{\textheight}%
}{0.5ex}}%
\stackon[1pt]{#1}{\tmpbox}%
}

\usepackage{stmaryrd} 
\usepackage[usenames]{color}
\usepackage{epsfig}
\usepackage{graphicx}
\usepackage{psfrag}

\graphicspath{{Figures/}}
\usepackage{float}
\floatstyle{boxed}
\newfloat{algorithm}{htb}{loa}
\floatname{algorithm}{Algorithm}
\usepackage{algorithm,algpseudocode}
\usepackage{caption}
\usepackage{subcaption}
\usepackage{color}
\usepackage{xcolor}
\usepackage{multirow}
\usepackage{natbib}
\modulolinenumbers[5]

\usepackage[left = 2cm, right=2.5cm, top=2cm, bottom =2cm]{geometry}

\begin{document}
\begin{frontmatter}

\title{ Integration of physics-informed operator learning and finite element method for parametric learning of partial differential equations }


\author{ Shahed Rezaei$^{1,*}$, Ahmad Moeineddin$^{2}$, Michael Kaliske$^{2}$, Markus Apel$^{1}$ }
\address{$^1$ACCESS e.V., Intzestr. 5, D-52072 Aachen, Germany}
\address{$^2$Institute for Structural Analysis, Technical University of Dresden, \\Georg-Schumann-Str. 7, D-01187 Dresden, Germany}
\address{$^*$ corresponding author: s.rezaei@access-technology.de}

\begin{abstract}
We present a method that employs physics-informed deep learning techniques for parametrically solving partial differential equations. The focus is on the steady-state heat equations within heterogeneous solids exhibiting significant phase contrast. Similar equations manifest in diverse applications like chemical diffusion, electrostatics, and Darcy flow. 
The neural network aims to establish the link between the complex thermal conductivity profiles and temperature distributions, as well as heat flux components within the microstructure, under fixed boundary conditions. A distinctive aspect is our independence from classical solvers like finite element methods for data. 
A noteworthy contribution lies in our novel approach to defining the loss function, based on the discretized weak form of the governing equation. This not only reduces the required order of derivatives but also eliminates the need for automatic differentiation in the construction of loss terms, accepting potential numerical errors from the chosen discretization method. As a result, the loss function in this work is an algebraic equation that significantly enhances training efficiency. We benchmark our methodology against the standard finite element method, demonstrating accurate yet faster predictions using the trained neural network for temperature and flux profiles. We also show higher accuracy by using the proposed method compared to purely data-driven approaches for unforeseen scenarios.
\end{abstract} 

\end{frontmatter}

\section{Introduction} 

Simulations of physical phenomena are recognized as the third pillar of science  \cite{Weinzierl2021}. 
While we focus on developing new models to accurately describe reality, a pressing and arguably more significant challenge lies in the efficient (i.e., rapid) and precise (i.e., high-accuracy) evaluation of these models. This challenge is particularly noteworthy in the context of developing digital twins or shadows \cite{BERGS202181}. 

Through the last decades, various numerical techniques have been developed for various applications. Numerical methods for solving model equations gain significant attention owing to their remarkable predictive capabilities. Among them, one can name finite difference, finite volume method, and finite element method \cite{Faroughi22, Liu2022}. Despite their great predictive abilities, these methods remain time-consuming in recent applications including multiphysics and coupled sets of nonlinear equations \cite{Ruan2022, REZAEI2023103758}. Another issue is that these methods cannot solve the equations in a parametric way. In other words, by any slight change in the input parameters, one has to repeat the calculations.

Machine learning (ML) techniques present a promising solution for overcoming the previously mentioned challenges, offering a considerable speed advantage over classical solvers once effectively trained \cite{Montans2023}. The efficiency of ML algorithms holds particular relevance in multi-scaling analysis (refer to \cite{fernandez2020, Peng2021} and references therein). While training time is a factor, it represents a one-time investment in contrast to the substantial speedup achieved. By comparing the training time with the evaluation time, one can assess the efficiency of employing ML techniques in various engineering applications (as demonstrated in comparisons studied in \cite{mianroodi2022lossless}). To ensure reliability, adequate training data encompassing relevant scenarios is essential. Moreover, integrating well-established physical constraints enhances prediction capabilities (see \cite{REZAEI2022PINN} and references therein). Investigating this aspect across diverse test cases and physical problems remains crucial for optimizing the application of physical laws in classical data-driven approaches. Our primary focus lies in exploring the potential of physics-driven deep learning methods within the computational engineering field, particularly in handling heterogeneous domains.

\color{black}
Considering the concept of physics-informed neural networks (PINNs) as introduced by \citet{RAISSI2019} and integrating physical laws into the final loss functions, the accuracy of predictions or network outcomes can be significantly improved. Additionally, in scenarios where the underlying physics of the problem is completely known and comprehensive, it becomes possible to train the neural network without any initial data \cite{REZAEI2022PINN, RAISSI2019, TORABIRAD2020109687}. This latter point is often referred to as unsupervised learning, as it does not require the solution to the problem beforehand. Comparable studies, such as the work by \citet{SIRIGNANO20181339}, propose methods like the deep Galerkin approach, which leverages deep neural networks to approximate solutions for high-dimensional partial differential equations.
\citet{Markidis2021} demonstrated that the standard version of PINNs can quickly converge for low-frequency components of the solution, while accurately capturing high-frequency components demands a longer training period.
In response to these shortcomings, extensions to the PINN framework have been proposed to enhance its performance. Some advancements are rooted in the domain decomposition method \cite{JAGTAP2020113028, jagtap2020extended, Moseley2023}, while others focus on novel architectures and adaptive weights to augment accuracy \cite{HAGHIGHAT2021, WANG21, McClenny22}. Recent investigations have introduced promising extensions involving the reduction of differential orders to construct loss functions and utilizing the energy form of equations \cite{REZAEI2022PINN, E2018, SAMANIEGO2020112790, FUHG2022110839}.

\color{black}
Despite the above progress, one main issue with solving partial differential equations using standard PINNs is that the solutions are still limited to specific boundary value problems. The latter is also the main issue with classical numerical solvers, as mentioned at the beginning of the introduction. Furthermore, in many engineering applications, the training time of the PINN, even with the most advanced optimizers and enhancements, cannot compete with those of more established numerical methods such as the finite element method. Nevertheless, thanks to the flexibility and interpolation power of neural networks, they have the potential to be extended and trained on a wider range of input parameters. Here, we review two main strategies for parametric learning of differential equations.

The first potential approach to generalize solutions obtained from neural networks is to employ transfer learning algorithms. This involves transferring pre-trained hyperparameters of a neural network to initialize the NN for the new domain \cite{Zhuang21, REZAEI2022PINN, GOSWAMI20TPF, XU2023115852, Harandi2023}.
Another strategy involves leveraging operator learning, which entails mapping two function spaces to each other. Some well-established methods for operator learning include but are not limited to, DeepOnet \cite{Lu2021} and Fourier Neural Operator \cite{li2021fourier}. For an in-depth exploration of this topic, refer to \cite{Faroughi2022, hildebrand2023comparison}.

\begin{figure}[H] 
  \centering
  \includegraphics[width=0.9\linewidth]{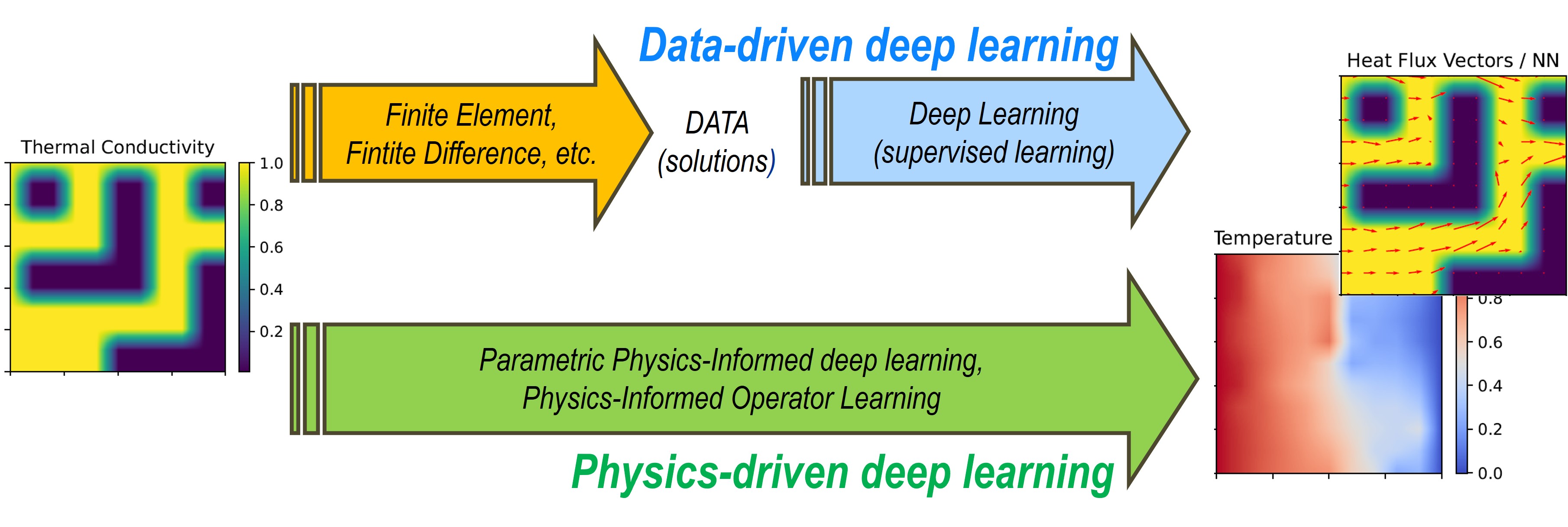}
  \caption{Utilizing physics-informed operator learning, one can enhance the classical data-driven approach and bypass the need for supervised learning for forward problems. }
  \label{fig:intro}
\end{figure}

The concept of PINN can also be integrated with operator learning (OL) to broaden the range of parameters for which the network can predict solutions. In \cite{Wang_Paris2021}, the authors utilized a physics-informed DeepONet framework to solve engineering problems. This idea is further illustrated in Fig.~\ref{fig:intro}, demonstrating how physics-driven operator learning can compete against traditional data-driven neural networks in efficiently and reliably mapping parametric input field(s) to solution field(s).
\color{black}

Several contributions utilizing physics-driven operator learning are outlined below. 
\citet{KORIC2023123809} investigated data-driven and physics-based DeepONets for solving heat conduction equations incorporating a parametric source term.
\citet{ZHU201956} utilized a convolutional encoder-decoder neural network to train physics-constrained deep learning models without labeled data.
\citet{GAO2021110079} introduced a novel physics-constrained CNN learning architecture based on elliptic coordinate mapping, facilitating coordinate transforms between irregular physical domains and regular reference domains.
\citet{ZHANG2021100220} introduced a physics-informed neural network tailored for analyzing digital materials, training the NN without labeled ground data using minimum energy criteria. 
\citet{LI2023116299} proposed a physics-informed operator neural network framework predicting dynamic responses of systems governed by gradient flows of free-energy functionals.
For achieving higher efficiency and potentially greater accuracy, \citet{kontolati2023learning} suggested mapping high-dimensional datasets to a low-dimensional latent space of salient features using suitable autoencoders and pre-trained decoders. \citet{ZHANG2023116214} presented encoder-decoder architectures to learn how to solve differential equations in weak form, capable of generalizing across domains, boundary conditions, and coefficients.

\color{black}
While physics-informed operator learning has shown intriguing and promising results, computing derivatives concerning the input variables via automatic differentiation is extremely time-consuming, especially when addressing higher-order partial differential equations. As a response, a recent trend in the literature aims to approximate derivatives more efficiently using classical numerical approaches. Employing this methodology transforms the loss function from a differential equation to an algebraic one. However, it is important to note that this approach may introduce discretization errors, influencing the network's outcomes.
\citet{Fuhg2023} introduced the deep convolutional Ritz method, a parametric PDE solver relying on minimizing energy functionals. Additionally, studies by \citet{REN2022114399}, \citet{ZHAO2023105516}, and \citet{ZHANG2023111919} explore physics-driven convolutional neural networks based on finite difference and volume methods.
In their work \cite{Hansen_2024}, authors proposed a framework for incorporating conservation constraints into neural networks, employing common discretization schemes like the trapezoidal rule for integral operators. Moreover, investigations by \citet{Phillips2023} showcase a novel approach leveraging artificial intelligence software libraries to replicate the process of solving partial differential equations using standard numerical methods like finite volume or finite element schemes.

\begin{figure}[H] 
  \centering
  \includegraphics[width=0.9\linewidth]{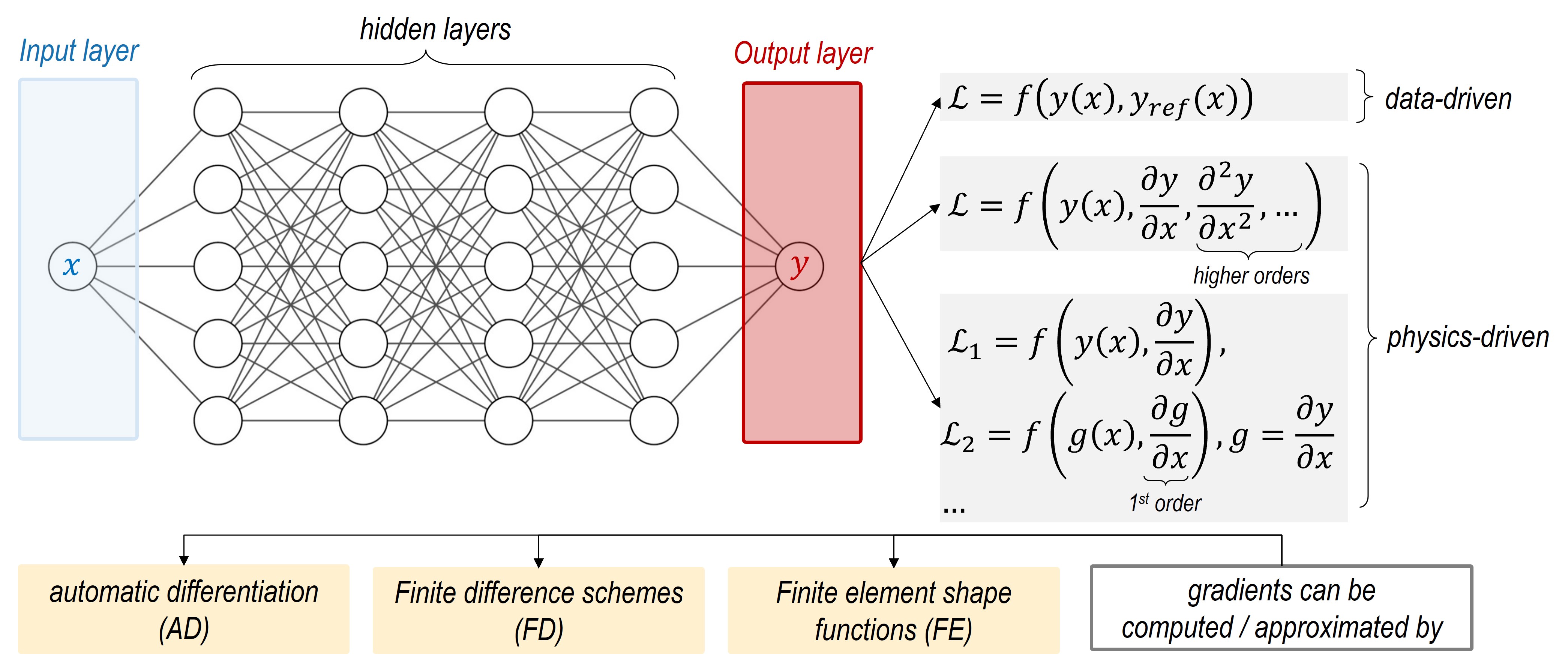}
  \caption{Difference between data and physics driven neural networks. The derivations for constructing differential equations can be approximated by different approaches. }
  \label{fig:motiv}
\end{figure}

Figure~\ref{fig:motiv} presents a comparison between data-driven and physics-driven neural networks, showcasing various residual design options. Additionally, the lower part of the figure illustrates the evaluation or approximation of gradients and integrals within the physical loss terms using different numerical schemes.
\color{black}
\color{black}

It is important to consider that operator or parametric learning, derived from available physical data obtained via numerical solvers, represents one possible approach (as depicted in the upper branch of Fig.~\ref{fig:intro}). For further exploration on this topic, refer to studies conducted by \citet{LU2022114778} and \citet{RASHID2023105444}, which provide comparative analyses of various neural operators across different applications.
\color{black}


The reviewed works represent significant contributions to both data-driven and physics-driven operator learning. However, each approach presents its potential drawbacks and advantages, and they still undergo development. This work's primary novel contribution lies in introducing a straightforward approach for physics-based operator learning, drawing inspiration from finite element discretization and residuals. Our proposed approach, coined as \underline{F}inite \underline{O}perator \underline{L}earning (FOL), offers flexibility and adaptability for various parametric inputs and outputs. Additionally, we will compare the outcomes of data-driven and physics-driven deep learning and discuss their respective accuracies and efficiencies, as well as their benefits and drawbacks.

After this introduction, in Section 2, we review the main governing equation of a steady-state thermal diffusion problem. Additionally, we explore the derivation of the discretized weak form of the PDE using the finite element technique.
Moving to Section 3, we focus on the application of deep learning to resolve the unknowns within the discretized weak form of the problem. Furthermore, we provide potential ideas on generating samples for training purposes.
In Section 4, we present the outcomes and findings of the conducted studies. Finally, the paper concludes with discussions on potential future developments.

\newpage
\section{Governing equation and its discretized weak form}
We are exploring a well-established problem of steady-state diffusion in a 2D setting and a heterogeneous solid. It is worth noting that the same derivation principles can be applied to investigate analogous fields such as electrostatics, electric conduction, or magnetostatic problems \cite{Guo2022}. Additionally, this approach extends to fields like chemical diffusion \cite{REZAEI2021104612} (under some assumptions) and Darcy flow \cite{li2023physicsinformed, YOU2022111536, Goswami2023}. In this context, the heat flux $\bm{q}$ is defined based on Fourier's law
\begin{align}
\label{Fourier}
\bm{q} = -k(x,y)\,\nabla T. 
\end{align} 
Here $k(x,y)$ is the phase-dependent heat conductivity coefficient. The governing equation for this problem is based on the heat balance which reads
\begin{align}
\label{StrongfromThermal}
\text{div}(\bm{q}) + Q &= 0~~~~~~ \text{in}~ \Omega, \\
\label{BCsthermalD}
T &= \bar{T}~~~~~\text{on}~ \Gamma_D, \\
\label{BCsthermalN}
\bm{q}\cdot \bm{n} = q_n &= \bar{q}~~~~~~\text{on}~ \Gamma_N.
\end{align} 
In the above relations, $Q$ is the heat source term. Moreover, the Dirichlet and Neumann boundary conditions are introduced in Eq.~\ref{BCsthermalD} and Eq.~\ref{BCsthermalN}, respectively. By introducing $\delta T$ as a test function and with integration by parts, the weak form of the steady-state diffusion problem reads
\begin{align}
\label{eq:weakformthermal}
\int_{\Omega}\,k(x,y)\,\nabla^T T\,\delta(\nabla T)~dV\,+\,\int_{\Gamma_N}\bar{q}~\delta T~dA\,-\int_{\Omega}\,Q\,\delta T~dV\,=\,0.
\end{align}
Utilizing the standard finite element method, the temperature field $T$, conductivity field $k$ as well as their first spatial derivatives, are approximated as
\begin{equation}
T= \sum {N_T}_i T_i =\boldsymbol{N}_{T} \boldsymbol T_e, \quad \nabla T =\sum {B_T}_i T_i =\boldsymbol{B}_{T}\boldsymbol T_e, \quad 
k= \sum {N_T}_i k_i =\boldsymbol{N}_{T} \boldsymbol k_e. 
\end{equation}
Here, $T_i$ and $k_i$ are the nodal values of the temperature and conductivity field of node $i$ of element $e$, respectively. Moreover, matrices $\boldsymbol N_T$ and $\boldsymbol B_T$ store the corresponding shape functions and their spatial derivatives. For a quadrilateral 2D element, they are written as
\begin{equation}
\boldsymbol N_T=\left[
\begin{matrix}
N_1    &\cdots & N_4 \\
\end{matrix}
\right],\quad 
\boldsymbol B_T=\left[
\begin{matrix}
N_{1,x} &\cdots & N_{4,x}  \\
N_{1,y} &\cdots & N_{4,y}\\
\end{matrix}
\right].
\end{equation}

\begin{equation}
\boldsymbol T^T_e=\left[
\begin{matrix}
T_1    &\cdots & T_4 \\
\end{matrix}
\right],\quad 
\boldsymbol k^T_e=\left[
\begin{matrix}
k_1    &\cdots & k_4
\end{matrix}
\right].
\end{equation}

The notation $N_{i,x}$ and $N_{i,y}$ represent the derivatives of the shape function $N_i$ with respect to the coordinates $x$ and $y$, respectively. To compute these derivatives, we utilize the Jacobian matrix $\boldsymbol J = \partial \boldsymbol X / \partial \boldsymbol \xi$, a standard procedure in any finite element subroutine \cite{bathe, hughes}. Here, $\boldsymbol X = [x, y]$ and $\boldsymbol \xi = [\xi, \eta]$ represent the physical and parent coordinate systems, respectively.

Going back to Eq.~\ref{eq:weakformthermal}, one can write the so-called \textit{discretized} version of the weak form for one general finite element as
\begin{align}
\label{eq:dis_residual}
\boldsymbol r_{eT} = \int_{\Omega_e} [\boldsymbol B_{T}]^T k~[\boldsymbol B_{T} \boldsymbol T_e] ~dV - \int_{\Omega_t}[\boldsymbol  N_{T}]^T k~[\boldsymbol B \boldsymbol T_e]^T~\boldsymbol  n~dS -\int_{\Omega} [\boldsymbol  N_{T}]^T  Q~dV.
\end{align}

\newpage
\section{Application of deep learning to solve parametric heat transfer equation}
This work is based on feed-forward neural networks.
Each neural network follows the standard structure, featuring a single input layer, potentially several hidden layers, and an output layer. Every layer is interconnected, transferring information to the next layer \cite{schmidhuber2015deep}. Within each layer, the neurons are not interconnected. Thus, we denote the information transfer from the $l-1$ layer to $l$ using the vector $\bm{z}^l$. Each component of vector $\bm{z}^l$ is computed by
\begin{equation}
\label{eq:NN_1}
    {z}^l_m = {a} (\sum_{n=1}^{N_l} w^l_{mn} {z}_n^{l-1} + b^l_{m} ),~~~l=1,\ldots,L. 
\end{equation}
In Eq.\,(\ref{eq:NN_1}), ${z}^{l-1}_n$, is the $n$-th neuron within the $l-1$-th layer. The component $w_{mn}$ shows the connection weight between the $n$-th neuron of the layer $l-1$ and the $m$-th neuron of the layer $l$. Every neuron in the $l$-th hidden layer owns a bias variable $b_m^l$. The number $N_I$ corresponds to the number of neurons in the $l$-th hidden layer. The total number of hidden layers is $L$. The letter $a$ stands for the activation function in each neuron. The activation function $a(\cdot)$ is usually non-linear. 

Building upon the concept of physics-informed neural networks (PINNs) proposed by \citet{RAISSI2019}, we construct the neural network's loss function based on the governing physical equations, alongside the initial and boundary conditions. In a standard PINN framework, the network's input comprises the locations of the collocation points. In the context of operator learning, we redefine these points as \textit{collocation fields}, which constitute randomly generated and admissible parametric spaces used to train the neural network. In our paper's context, these collocation fields represent possible choices for the conductivity parameter within the spatial domain, denoted as $k(x,y)$.

\begin{figure}[H] 
  \centering
  \includegraphics[width=0.99\linewidth]{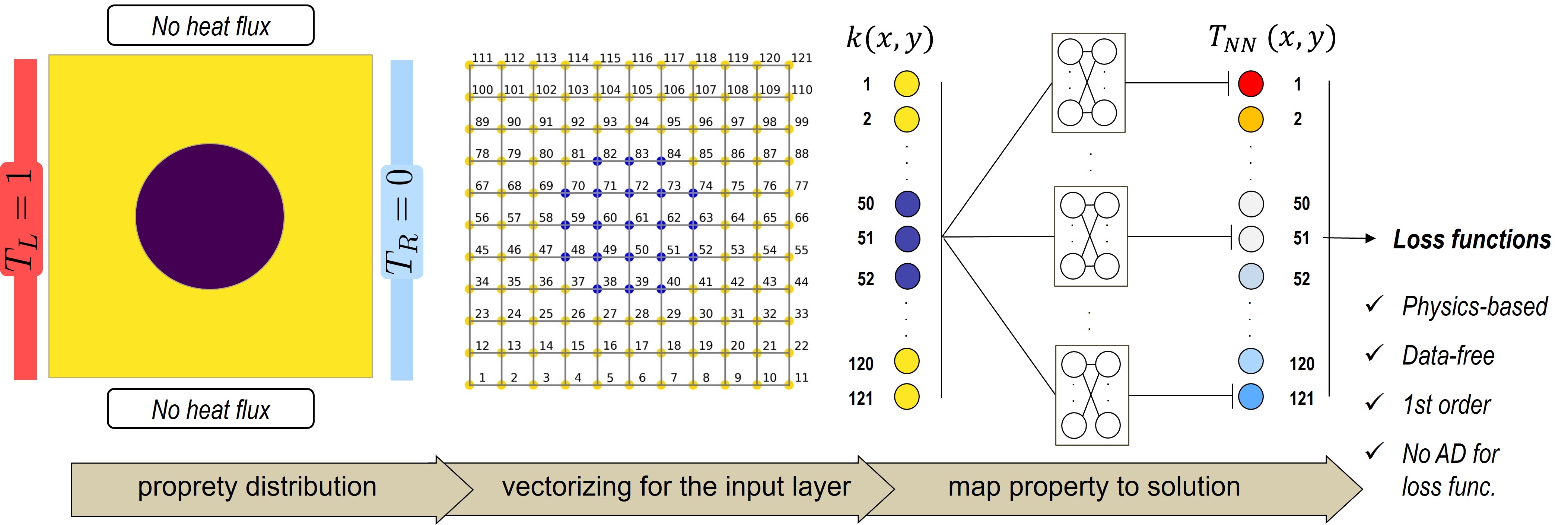}
  \caption{ Network architecture for finite operator learning, where information about the input parameter at each grid point goes in and the unknown is evaluated utilizing a series of separate deep neural networks. }
  \label{fig:NN_idea}
\end{figure}

The output layer $\bm{Y}={T_i}$ consists of the components representing the temperature field at each node. We propose employing separate fully connected feed-forward neural networks for each output variable (as depicted on the right-hand side of Fig.~\ref{fig:NN_idea}). The results obtained from these neural network models are
\begin{align}
\label{eq:in_out}  
    T_{i} = \mathcal{N}_{i} (\bm{X}; \bm{\theta}_i),~~~\bm{X}=\{k_j\},~~~\bm{\theta}_i=\{\bm{W}_i,\bm{b}_i\},~~~i,j = 1 \cdots N.
\end{align}
Here, the trainable parameters of the $i$-th network are denoted by $\bm{\theta}_i$. Moreover, $\bm{W}_i$ and $\bm{b}_i$ encompass all the weights and biases within the $i$-th neural network. Additionally, $k_j$ represents the conductivity value at node $j$. The number of nodes is denoted by $N$, set to $121$ in this paper.

Next, we introduce the loss function and propose a new method to evaluate it accurately without using automatic differentiation. First, we assume no heat source term, i.e., $Q=0$. Second, we consider the upper and right edges as isolated, resulting in a vanishing normal heat flux at the Neumann boundaries. The temperature values at the left and right edges are fixed at $1.0$ and $0.0$, respectively (refer to Fig.~\ref{fig:NN_idea}). 

The total loss term $\mathcal{L}_{\text{tot}}$ integrates both the elemental energy form of the governing equation $\mathcal{L}_{\text{en,e}}$ and the Dirichlet boundary terms $\mathcal{L}_{\text{db,i}}$. To minimize the complexity of derivations, we directly employ the discretized energy form of the heat equation, akin to the weak form in Eq.~\ref{eq:dis_residual}. The integrals are accurately computed using Gaussian integration, resulting in
\begin{align}
\label{eq:loss_tot}
\mathcal{L}_{tot} &= \sum_{e=1}^{n_{el}}\mathcal{L}_{en,e}(\boldsymbol \theta) + \sum_{i=1}^{n_{db}} \mathcal{L}_{db,i}(\boldsymbol \theta), \\
\label{eq:loss_ene}
\mathcal{L}_{en,e} &= \boldsymbol T^T_e(\boldsymbol \theta) \left[ \boldsymbol K_e \boldsymbol T_e(\boldsymbol \theta) \right], \\
\label{eq:loss_db}
\mathcal{L}_{db,i} &= \dfrac{1}{n_{db}} |T_i(\boldsymbol \theta) - T_{i,db}|, \\
\boldsymbol K_{e} &= \sum_{n=1}^{n_{int}} \dfrac{w_n}{2}~\text{det}(\boldsymbol J)~[\boldsymbol B_{T}(\boldsymbol \xi_n)]^T \boldsymbol C_e(\boldsymbol \xi_n) \boldsymbol B_{T}(\boldsymbol \xi_n), \\
\boldsymbol{C}_e(\boldsymbol \xi_n) &= \left[
\begin{matrix}
\boldsymbol{N}_{T}(\boldsymbol \xi_n) \boldsymbol k_e & 0  \\
0 & \boldsymbol{N}_{T}(\boldsymbol \xi_n) \boldsymbol k_e\\
\end{matrix}
\right].
\end{align}
In the above relation, $n_{\text{int}}=4$ represents the number of Gaussian integration points. Additionally, $\boldsymbol \xi_n$ and $w_n$ denote the coordinates and weighting of the $n$-th integration point. $\text{det}(\boldsymbol J)$ stands for the determinant of the Jacobian matrix. It is worth mentioning that this determinant remains constant for parallelogram-shaped elements, eliminating the need to evaluate this term at each integration point.
\begin{figure}[H] 
  \centering
  \includegraphics[width=0.99\linewidth]{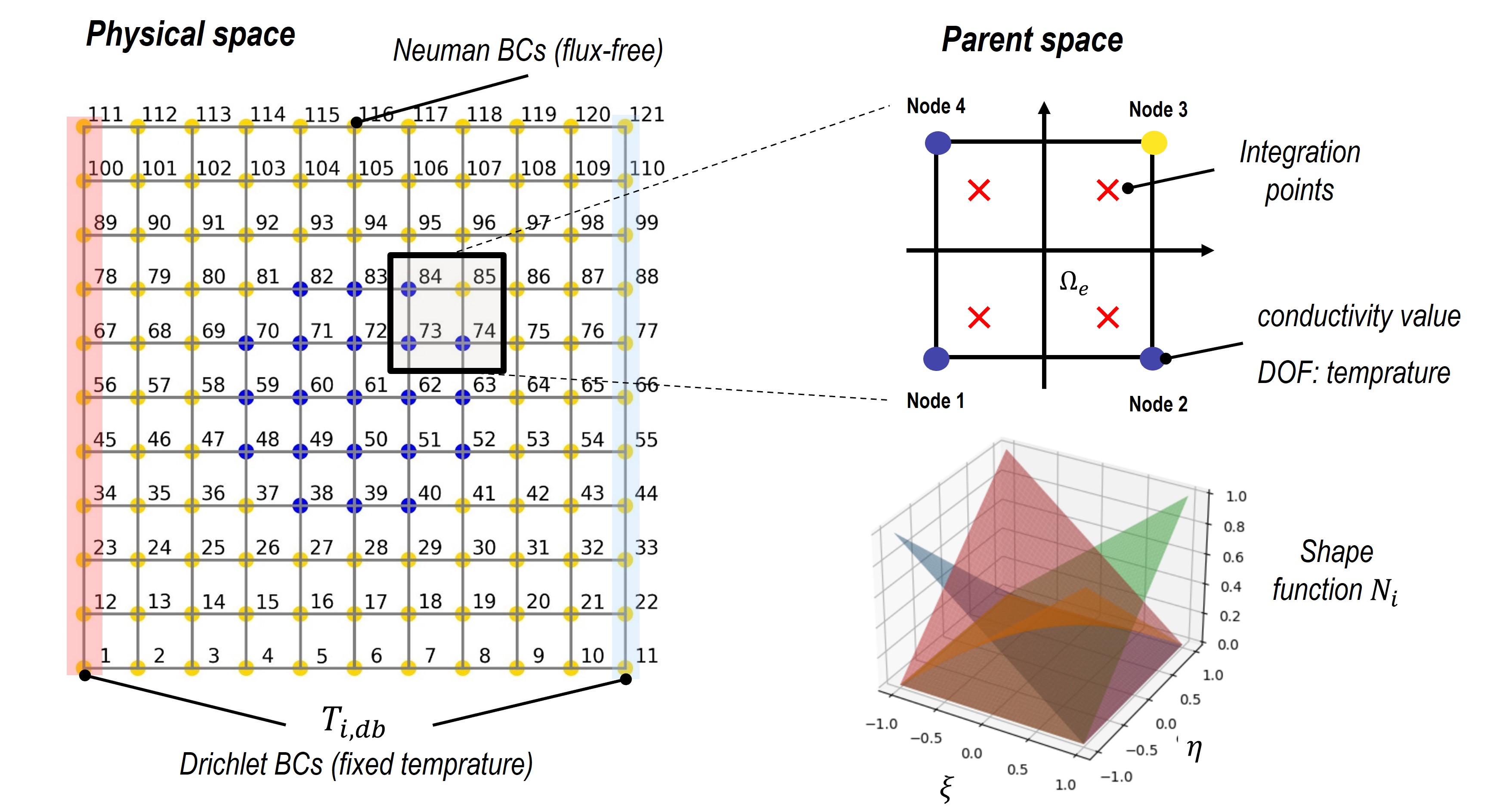}
  \caption{The physical space and the parent space. Gaussian integration and shape functions are executed and defined specifically within the parent space. }
  \label{fig:FE_int}
\end{figure}
As depicted in the upper right section of Fig.~\ref{fig:FE_int}, we utilize the positions of the standard 4 Gaussian points within the parent space to compute the energy integral.
In Eq.\ref{eq:loss_db}, $n_{\text{db}}=22$ denotes the number of nodes along the Dirichlet boundaries (i.e., the left and right edges). Additionally, $T_{i,\text{db}}$ represents the given (known) temperature at node $i$. The final equation for the loss function is summarized below, subsequently implemented in Python using the Sciann package \cite{SciANN}
\begin{align}
\label{eq:loss_sum}
\boxed{
\mathcal{L}_{tot} =  
\underbrace{
\lambda_e
\sum_{e=1}^{n_{el}}\boldsymbol T^T_e \left[ \boldsymbol \sum_{n=1}^{n_{int}} ~(\boldsymbol N_T \boldsymbol k_e)\boldsymbol B_{T}^T \boldsymbol B_{T} \right] \boldsymbol T_e 
}_{\text{energy + Neumann BCs}} 
+
\quad \lambda_b\underbrace{\sum_{i=1}^{n_{db}} |T_i - T_{i,db}|}_{\text{Drichlet BCs}}.
}
\end{align}
In the above equation, our specific focus lies on quadrilateral elements where the determinant of the Jacobian matrix remains constant. Additionally, we solely utilized 4 integration points with weightings $w_n=1$. We define $\lambda_e=\dfrac{w_n}{2}\text{det}(\boldsymbol J)$ and $\lambda_b=\dfrac{\Lambda}{n_{\text{db}}}$. Furthermore, in Eq.\ref{eq:loss_sum}, $\lambda_b$ represents the weighting of the boundary terms. As we will elaborate later, we intentionally increase this term to efficiently enforce the boundary conditions during training.

The final loss term is minimized concerning a batch of collocation fields. This mathematical optimization problem is expressed as  
\begin{align}
\label{minimize}
\bm{\theta}^* = \arg \min_{\bm{\theta}} \mathcal{L}_{tot}(\bm{X}; \bm{\theta}),
\end{align}
where $\bm{\theta}^*$ represents the optimal trainable parameters of the network (weights and biases).

In this work, we utilized the Adam optimizer \cite{kingma2014adam}. The training predominantly occurred on personal laptops without relying on GPU acceleration, as the computations were not computationally intensive. Several parameters and hyperparameters significantly affect the network’s performance, including the activation function, the number of epochs, the batch size, the number of hidden layers, and the number of neurons within each layer. The impact of these parameters is further explored in the Results section.
\\ \\
\noindent
\textbf{Remark 1} It is also beneficial to directly enforce the Dirichlet boundary terms using a hard boundary approach \cite{Harandi2023}. In this scenario, the last term in Eq.~\ref{eq:loss_sum} vanishes. Implementing such a modification is straightforward in the current approach due to the clear identification of nodes with known temperatures. Nonetheless, we aim to demonstrate that even by penalizing the boundary term, successful training is achievable. One reason is to facilitate future advancements, allowing for parametric alterations in the boundary values during training to generalize the network solution.
\\ \\
\textbf{Remark 2} As described in Eq.\ref{eq:in_out} and depicted in Fig.~\ref{fig:NN_idea}, we recommend utilizing separate neural networks for predicting nodal temperature values. Later, we will demonstrate how profoundly this aspect influences the result's quality. Moreover, adopting this approach provides better insights into the necessary architecture for the neural network.
\\ \\
\textbf{Remark 3} In Eq.~\ref{eq:loss_sum}, there is no involvement of automatic differentiation, and all the derivatives are approximated using the shape functions from the finite element method. In simpler terms, the loss function in this work is an algebraic equation. It is worth noting that, we approximate the derivatives, which may not be as accurate as the automatic differentiation technique.
\\

The last preparatory step before training the neural network involves generating samples for training. Since the training process is entirely unsupervised and does not require knowledge of the solution, it is sufficient to create enough samples to cover the domain of interest. These samples are then used to train the neural network. In this study, we confine ourselves to a two-phase material where the property (i.e., thermal conductivity) of each phase remains fixed. However, the microstructure's morphology is altered, allowing for arbitrary shapes and volume fraction ratios between the two phases. It is noteworthy that for a selected reduced grid (with 121 nodes) and considering a two-phase material, there can be approximately \(2^{121} \approx 2.6 \times 10^{36}\) independent samples. Creating and feeding the network with such a vast number of samples for further training is almost unfeasible. Hence, the user should determine the relevant parameter space specific to each problem, thereby preventing unnecessary complications.

For the time being, we suggest the following strategy for creating samples and we shall explore the impact of the sample size on the network predictions in the results section. For reader convenience and to ensure reproducibility of results, the main steps are outlined below:
\begin{itemize}
\item Choose a set of random center points based on a user-defined number.
    \item Choose random values for the inner radius, outer radius, and rotation angle of the ellipse; the ranges for these values are all user-defined.
    \item Assign a (low) conductivity value for the grid nodes inside the defined region.
\end{itemize}

\begin{figure}[H] 
  \centering
  \includegraphics[width=0.85\linewidth]{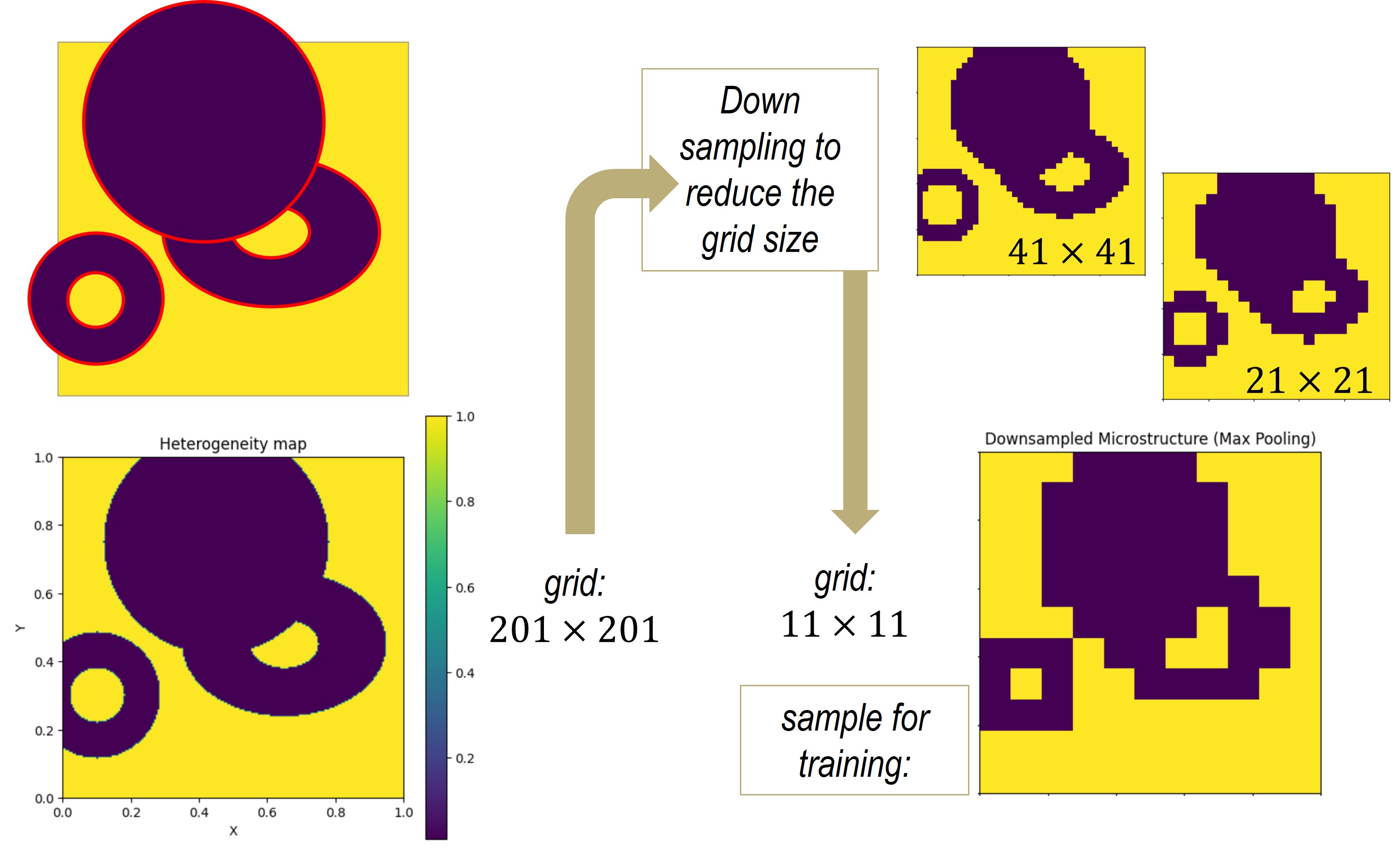}
  \caption{Downsampling the grid size involves using convolutional layers combined with max-pooling, resulting in a reduced space of $11 \times 11 = 121$ grid points. The training will be conducted based on an 11 by 11 grid size, resembling the concept of a latent space.}
  \label{fig:shapes}
\end{figure}

The described procedure is implemented into Python, generating 4000 samples for a grid of size $N$ by $N$, where $N=11$. The choice of $N=11$ aims to minimize training time and reduce the neural network's size. This approach can also be achieved by downsampling a high-resolution image of the microstructure (see Fig.~\ref{fig:shapes}). This transformation can be viewed as a \textit{latent space}, a lower-dimensional representation that captures essential features or patterns of the input data. In future studies, it is worth investigating the ideal size of the grid for training. 

\begin{figure}[H] 
  \centering
  \includegraphics[width=0.99\linewidth]{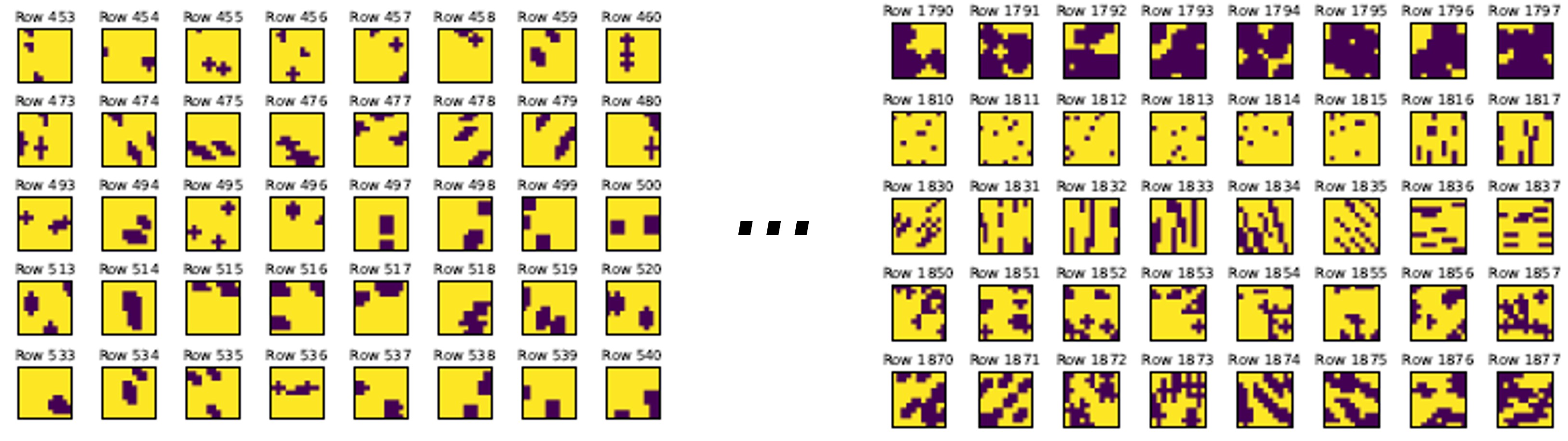}
  \caption{Examples showcasing created morphologies for a two-phase microstructure.}
  \label{fig:samples}
\end{figure}

\begin{figure}[H] 
  \centering
  \includegraphics[width=0.99\linewidth]{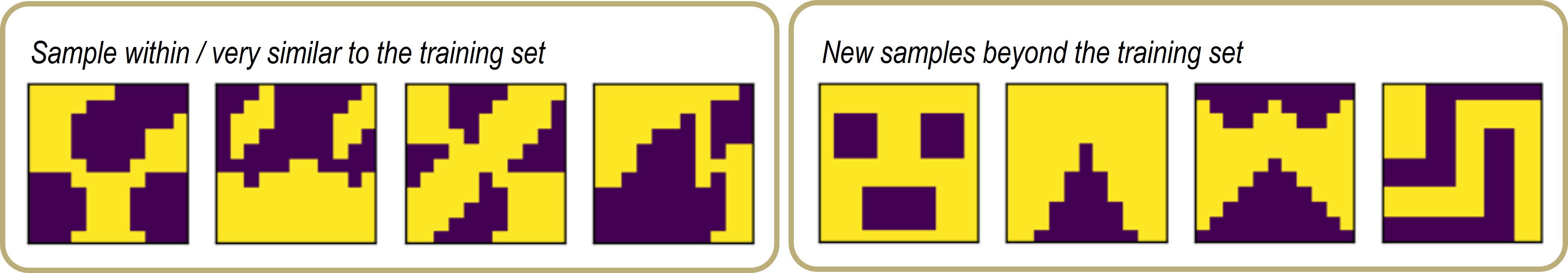}
  \caption{Unseen samples used for testing the neural network after training. }
  \label{fig:test}
\end{figure}

\begin{figure}[H] 
  \centering
  \includegraphics[width=0.9\linewidth]{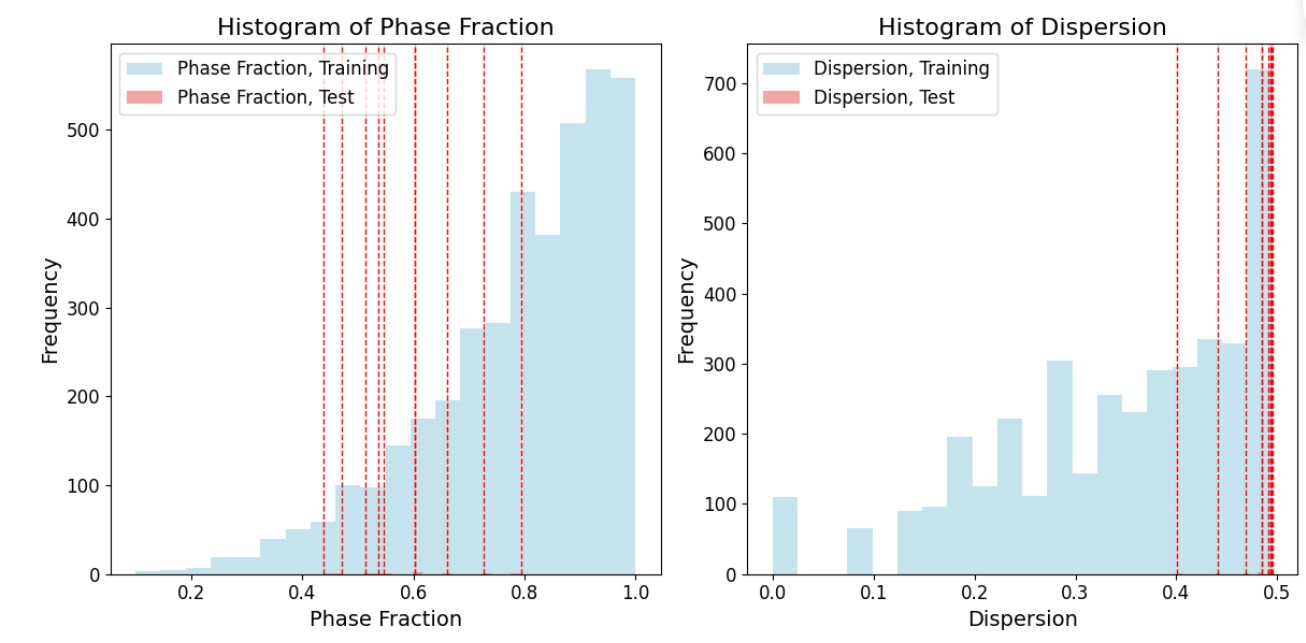}
  \caption{Histogram of the phase fraction as well as dispersion values of the created samples. The red lines show the location of the test samples. }
  \label{fig:hist}
\end{figure}

\newpage
A subset of the generated samples is presented in Fig.~\ref{fig:samples}. Additionally, to assess the NN's performance, 8 additional samples are chosen for testing (see Fig.~\ref{fig:test}). The first 4 samples resemble the shapes of the generated ones. As for the last 4 test samples, they are entirely novel and deliberately selected for their unique symmetrical aspects, ensuring they are beyond the scope of the training dataset. Furthermore, to provide a comprehensive view of the created samples, the volume fraction and dispersion value are evaluated for each morphology. Corresponding histograms are depicted in Fig.~\ref{fig:hist}, where the red lines indicate the positions of the test samples.

\newpage
\section{Results}
The algorithms developed in this study are implemented using the SciANN package \cite{SciANN}, and the methodology can be adapted to other programming platforms. A summary of the network's (hyper)parameters is presented in Table~\ref{tab:NN_para}. Note that whenever multiple options for a parameter are introduced, we examine the influence of each parameter on the results.
\begin{table}[H]
\centering
\caption{Summary of the network parameters.}  
\label{tab:NN_para}
\begin{footnotesize}
\begin{tabular}{ l l }
\hline
Parameter                          &  Value    \\
\hline
Inputs, Outputs                  &  $\{T_i\}$, $\{k_i\}$ \\ 
Activation function                &  tanh, swish, sigmoid, linear \\ 
Number of layers and neurons for each sub network ($L$, $N_l$)  &  (2, 10) \\
Optimizer                         &  Adam \\ 
Number of initial samples                         &  $2000$, $4000$ \\ 
Batch size                         &  $50$, $100$ \\ 
(Learning rate $\alpha$, number of epochs)  &  $(10^{-3},5000)$ \\ 
\hline
\end{tabular}
\end{footnotesize}
\end{table}

The material parameters listed in Table \ref{tab:par} are selected to maintain the temperature values within the range of $0$ to $1$. However, alternative parameters and boundary values can be chosen, and the model variables can be normalized to ensure they fall within the range of $-1$ to $1$. 
\begin{table}[H]
\centering
\caption{Model input parameters for the thermal problem.}  
\label{tab:par}
\begin{footnotesize}
\begin{tabular}{ l l }
\hline
      &  heat conductivity value / unit    \\
\hline
Phase 1  ($k_{\text{mat}}$)  & $1.0$~W/mK  \\
Phase 2 ($k_{\text{inc}}$)  & $0.01$~W/mK  \\
Applied temperature ($T_L$,$T_R$)  & ($1.0$~K, $0.0$~K)\\
\hline
\end{tabular}
\end{footnotesize}
\end{table} 

\subsection{Evoltion of loss function and network prediction} 

\begin{figure}[H] 
  \centering
  \includegraphics[width=0.7\linewidth]{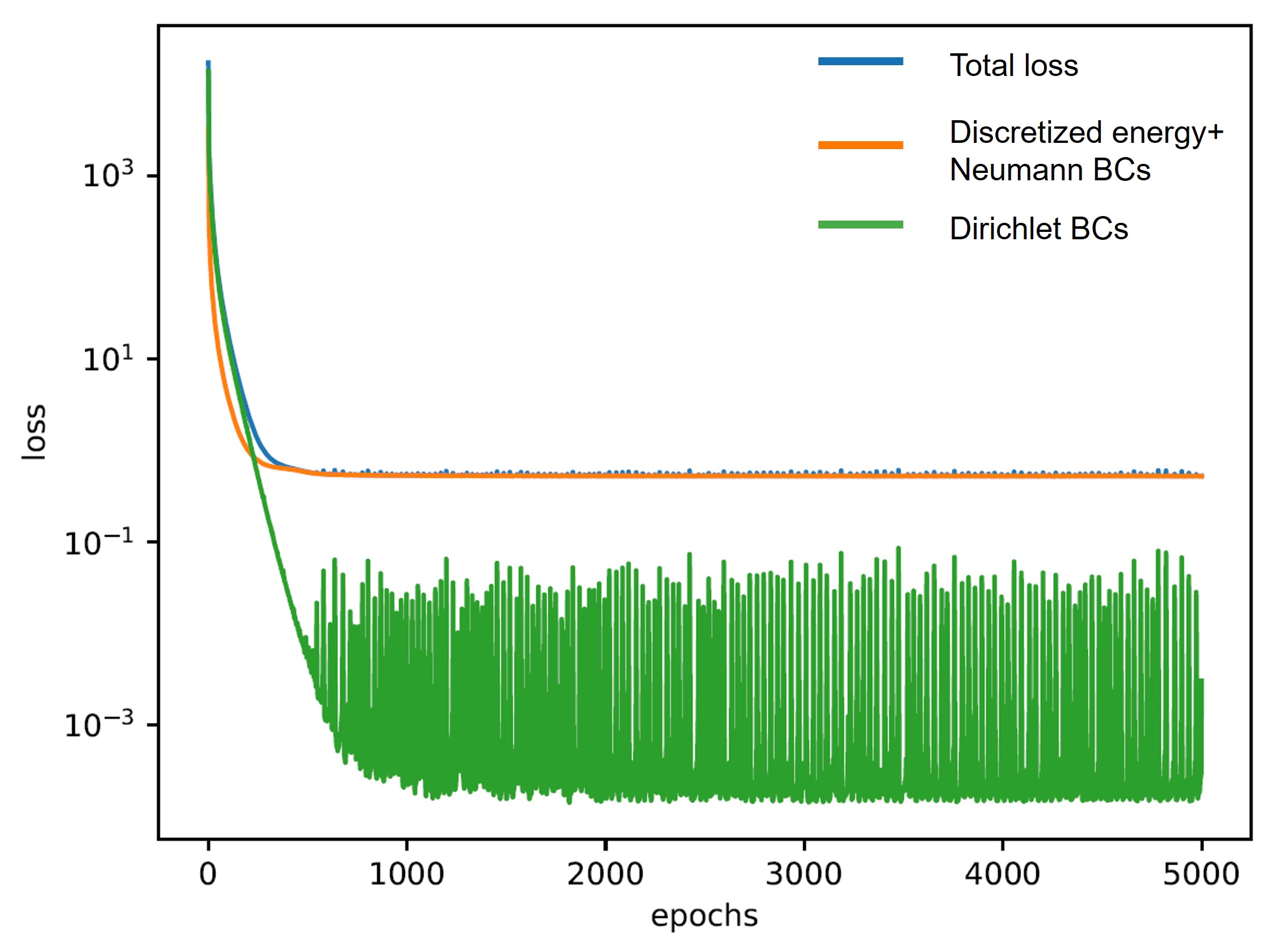}
  \caption{Evolution of loss terms for parametric learning of the steady-state thermal problem.}
  \label{fig:loss}
\end{figure}

The evolution of each loss term is shown in Fig.~\ref{fig:loss} as a function of epochs.  

The collocation domains for training are established as discussed in the previous section, comprising 4000 randomly generated two-phase microstructures on an 11 by 11 grid (see also Fig.~\ref{fig:samples}). In the results presented in Fig.~\ref{fig:loss}, we employ the "tanh" activation function, while the remaining parameters align with those in Table~\ref{tab:NN_para}. All the loss functions decay simultaneously, and we did not observe any significant improvement after $5000$ epochs. This consistent response was observed across various hyperparameter studies for all the loss terms. For each epoch, we required approximately 5 seconds to train the neural network with the specified hyperparameters on the Apple M2 platform. Further comparisons regarding the training and evaluation costs are detailed in Table \ref{tab:network_cost}.

For the loss term related to Dirichlet boundary conditions, we assigned higher weightings to expedite the satisfaction of the boundary terms. Following a systematic study that began with equal weightings for both loss terms, we selected $\Lambda=10~n_{nb}$, indicating $\lambda_b=10$. Choosing equal or lower weightings resulted in less accurate predictions within the same number of epochs, as the boundary terms were not fully satisfied. Opting for higher values did not significantly improve the results. In future studies, more advanced loss balancing techniques, as explored in \cite{WANG2022why, chen2018gradnorm}, can be employed to avoid the trial-and-error procedure observed in this study. 

\begin{figure}[H] 
  \centering
  \includegraphics[width=1.05\linewidth]{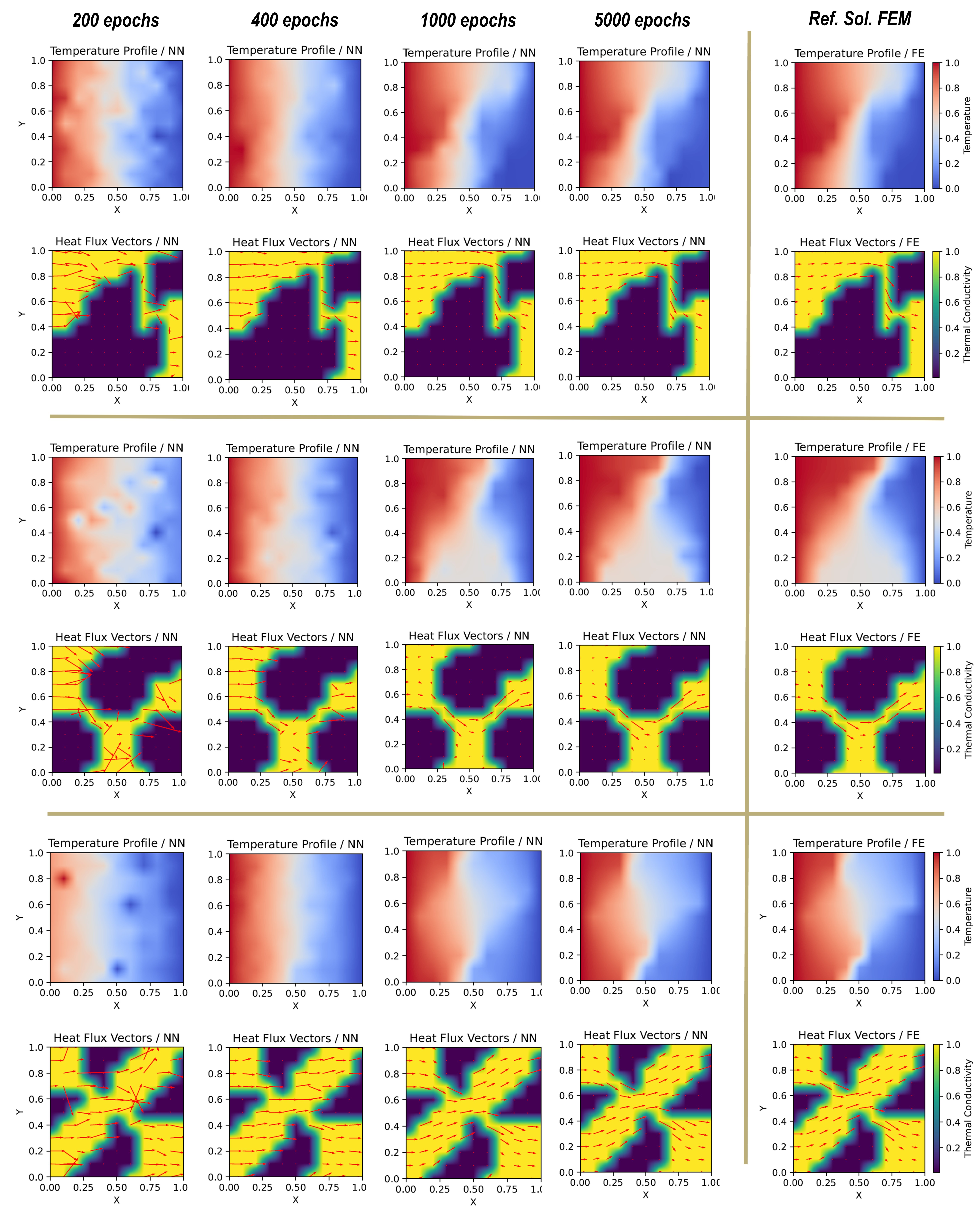}
  \caption{Predictions of the trained FOL for different numbers of epochs.}
  \label{fig:evol_1}
\end{figure}

\begin{figure}[H] 
  \centering
  \includegraphics[width=1.05\linewidth]{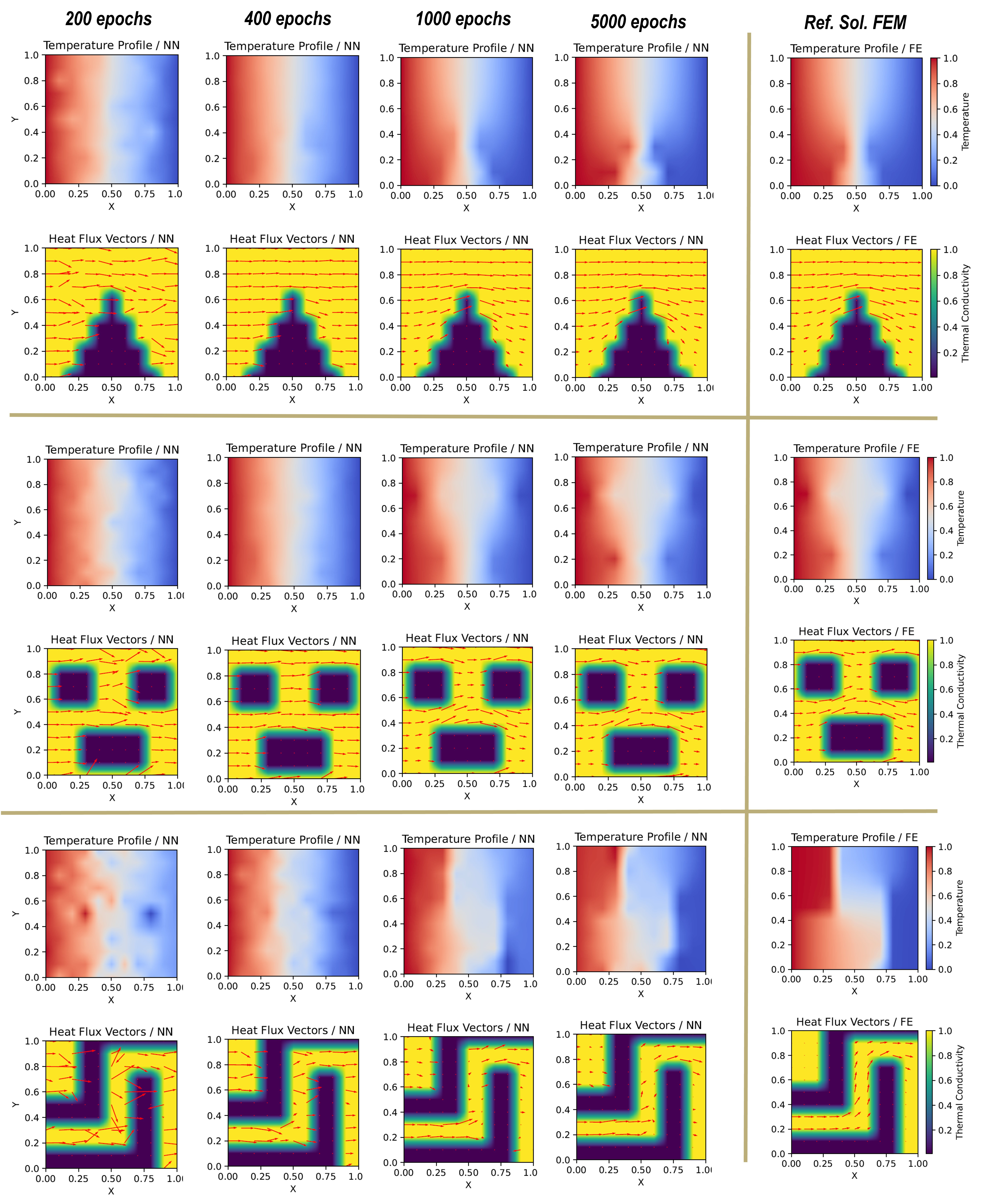}
  \caption{Predictions of the trained FOL for different numbers of epochs.}
  \label{fig:evol_2}
\end{figure}

In Figures \ref{fig:evol_1} and \ref{fig:evol_2}, the predictions of the same neural network for various sample test morphologies (i.e., conductivity maps) are depicted across different numbers of epochs, presented in separate columns. The last column displays the reference solutions obtained by the finite element method, utilizing the same number of elements (10 by 10 elements or 11 by 11 nodes). It is important to note that post-training, the results are assessed on finer grid points ($165 \times 165$) using simple shape function interpolations, which lend a smoother appearance to the results. For conciseness, all temperature values are denoted in Kelvin $[K]$, heat flux is represented in $[W/m^2]$, and thermal conductivity is expressed in $[W/mK]$.

As anticipated from Fig.~\ref{fig:loss}, the NN's outcomes are unreliable before 1000 epochs. Surprisingly, after 1000 epochs, the NN demonstrates an acceptable level of training. However, continuing the training up to 5000 epochs marginally improved the overall results. This improvement is particularly noticeable in sharp discontinuities near the phase boundaries (as evident in the first row of Fig.\ref{fig:evol_2}). It is noteworthy that the results showcased in Figures \ref{fig:evol_1} and \ref{fig:evol_2} correspond to test cases where these morphologies were unseen by the NN. Similar trends were observed for other test cases within or outside the training set, although not presented here to maintain conciseness in the results section. More comprehensive comparisons and quantitative error analyses are provided in subsequent sections.
\\ \\
\textbf{Remark 4} Rather than relying on finite element interpolation functions, which do not add any additional information and merely smooth the results, an alternative approach involves employing a pre-trained decoder network. This network can map the solution space from a coarse grid to a finer grid. Similar ideas have been explored in \cite{kontolati2023learning}.
\\ \\
\textbf{Remark 5} We exclusively predict temperature profiles in the current approach. Consequently, the flux vector, as presented in subsequent studies, is derived via local numerical differentiation of the temperature. However, an extension of this approach could involve a mixed formulation, predicting the flux values as additional outputs for the network to further enhance predictions, as explored in works such as \cite{Faroughi22, REZAEI2022PINN, Harandi2023}.

\newpage
\subsection{Comparison of physics-driven and data-driven deep learning models}
\begin{figure}[H] 
  \centering
  \includegraphics[width=0.75\linewidth]{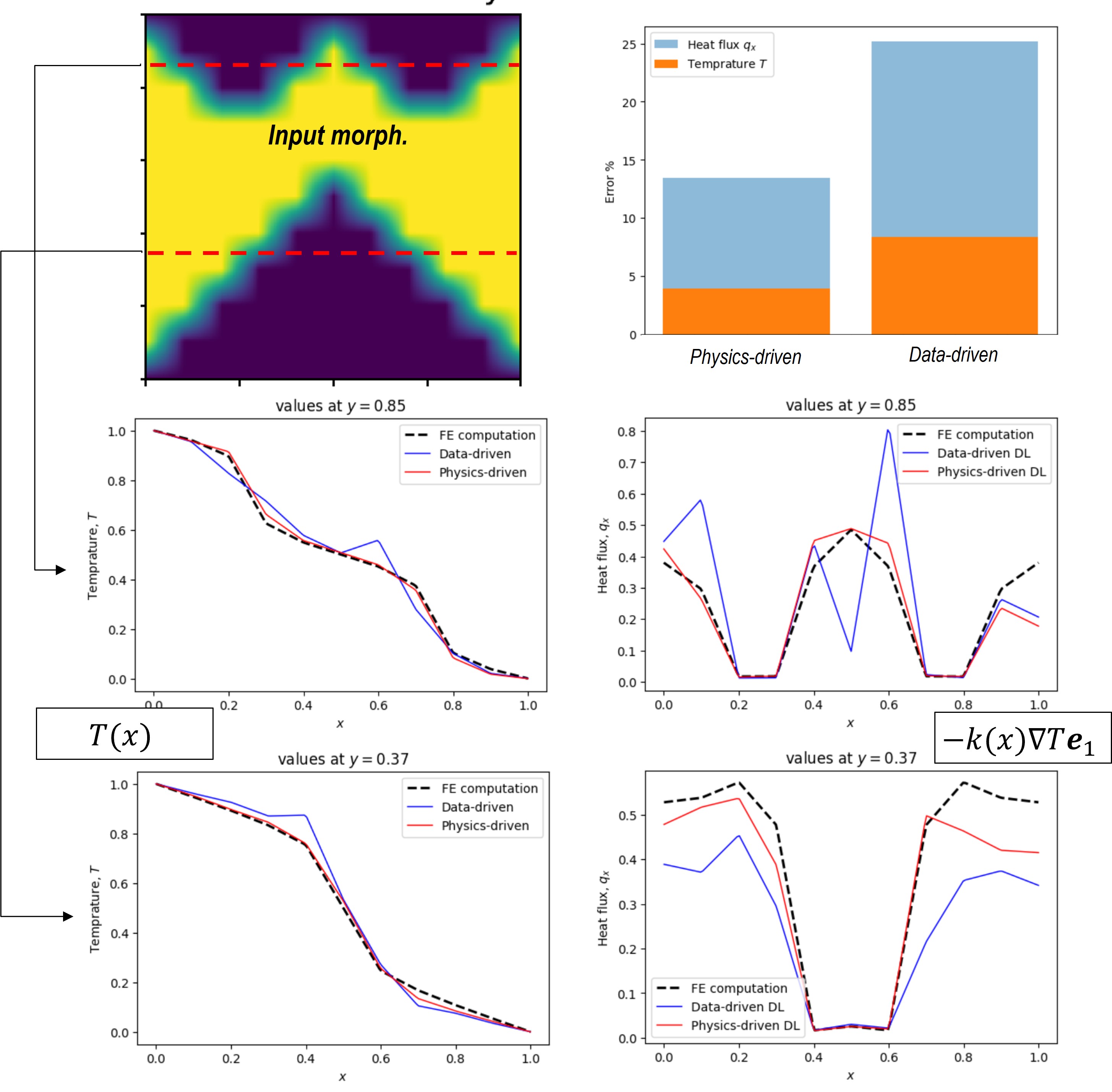}
  \caption{Comparison of data-driven and physics-driven approaches regarding the temperature and flux profiles.}
  \label{fig:err_dd}
\end{figure}

\begin{figure}[H] 
  \centering
  \includegraphics[width=0.99\linewidth]{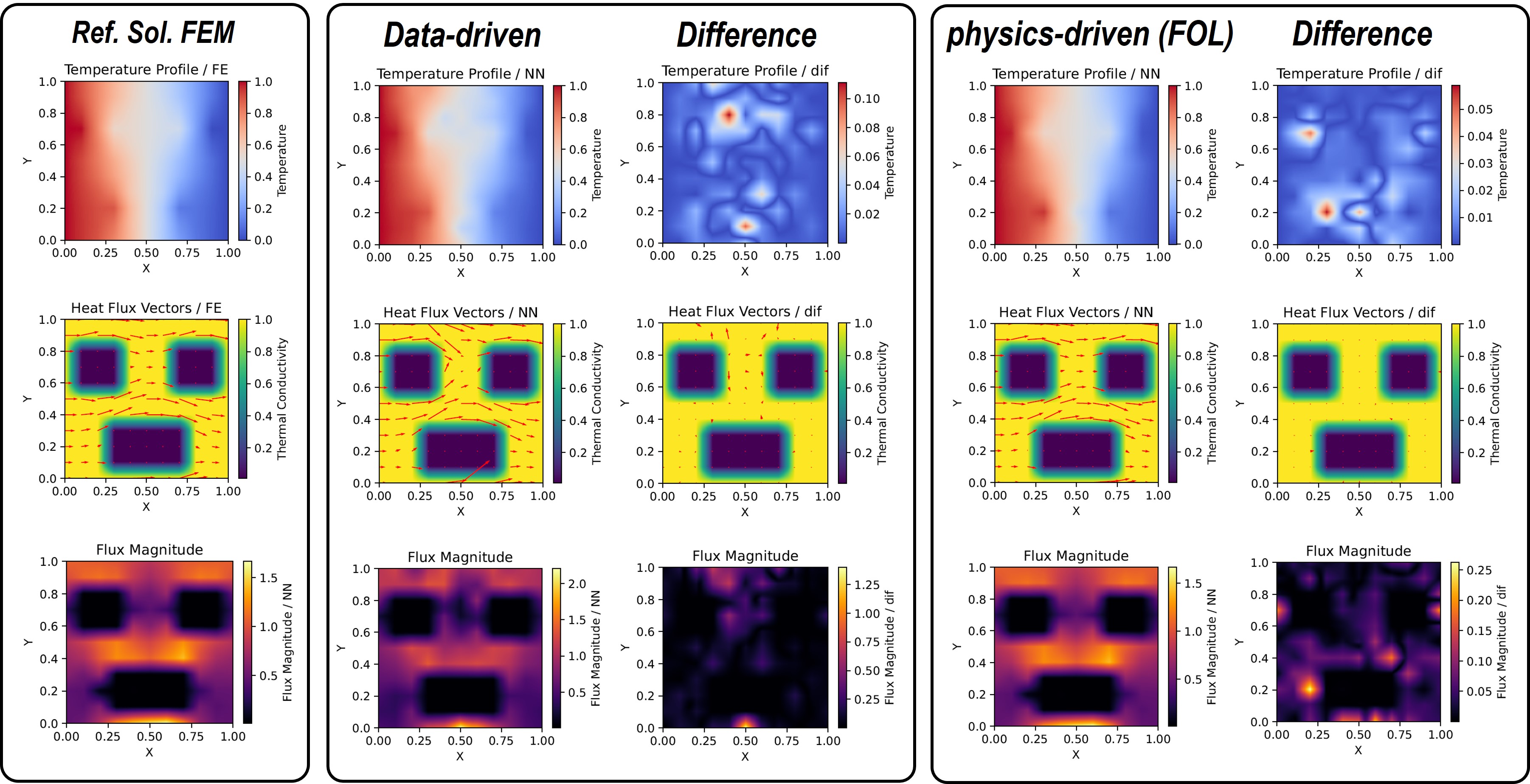}
  \caption{Comparison of data-driven and physics-driven approaches for unseen test cases.}
  \label{fig:DD_1}
\end{figure}

\begin{figure}[H] 
  \centering
  \includegraphics[width=0.99\linewidth]{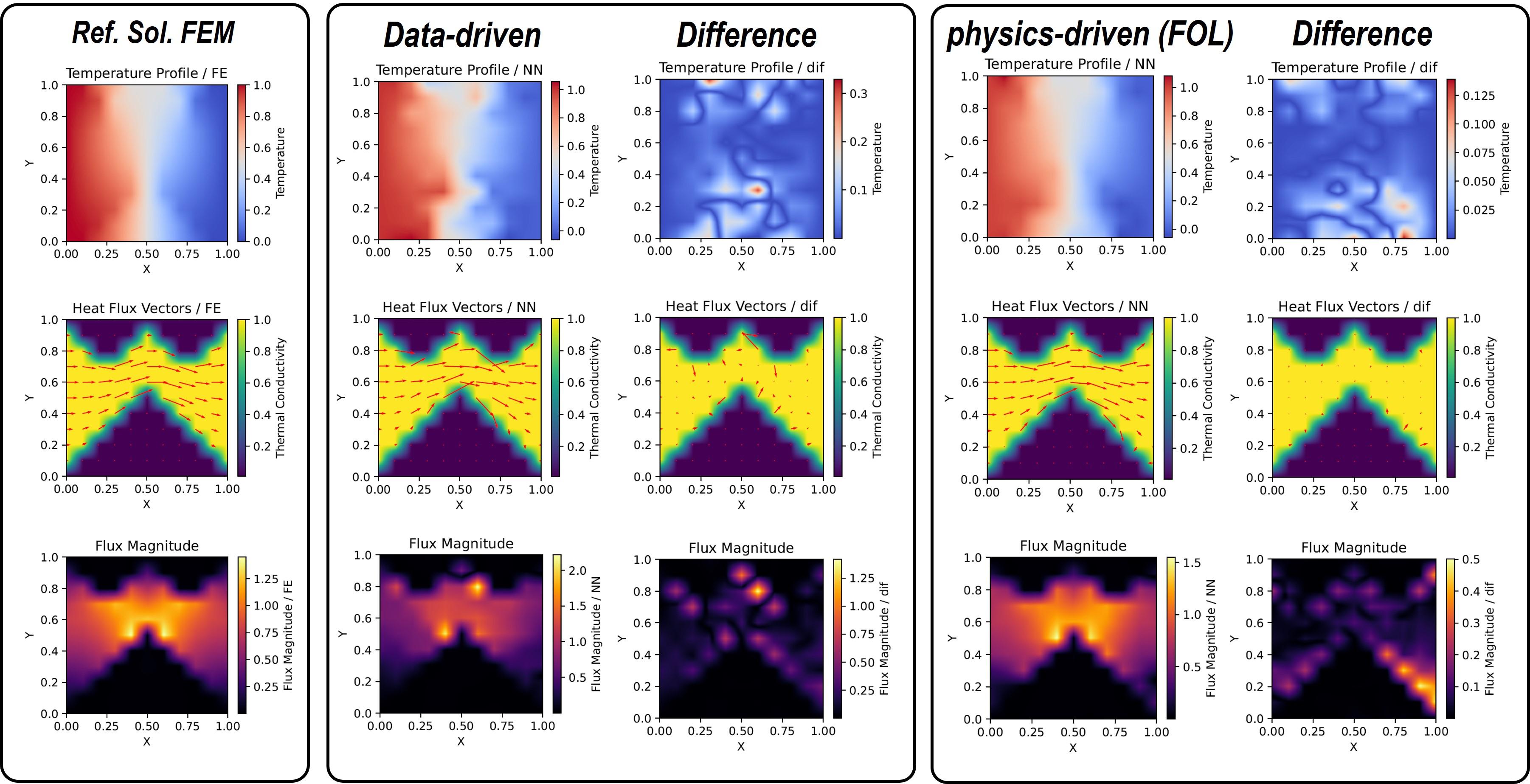}
  \caption{Comparison of data-driven and physics-driven approaches for unseen test cases.}
  \label{fig:DD_2}
\end{figure}

In this section, we aim to compare the results obtained from the physics-driven neural network model with those derived from a network trained through supervised learning using data obtained from the finite element method. Essentially, we seek to demonstrate the implications of using traditional numerical models, gathering their data, and subsequently training a neural network based on this available solution. This latter approach, termed data-driven, is widely pursued by many researchers, as briefly reviewed in the introduction section. To conduct this comparison, we utilize the same set of 4000 sample data, perform finite element calculations for each, and store the associated solutions. To ensure a fair comparison, we employ an identical network architecture to train the data-driven network. The comparison results are illustrated in Fig.~\ref{fig:err_dd} for a specific test case. Clearly, for an unseen test case, the physics-driven method provides more accurate predictions for temperature and flux distribution. The error in this study is based on $Err = \frac{\sqrt{\sum_i (T_{NN}(i)-T_{FE}(i))^2}}{\sqrt{\sum_i (T_{FE}(i))^2}}\times 100$.  
The observations and conclusions mentioned above are not confined to a single test case or the chosen cross sections; they manifest consistently across other unseen cases and regions as well (refer to Figures \ref{fig:DD_1} and \ref{fig:DD_2}). This significantly underscores the appeal of the physics-driven model for two primary reasons: 1) It does not rely on labeled data for training, and 2) it delivers more accurate predictions for unseen cases. However, it is crucial to emphasize a few points: 1) Although the training cost is a one-time investment, we should report that the supervised training requires a lower training cost (approximately 10 times less, on average) compared to the physics-based approach. 2) The data-driven model exhibits slightly better predictions for cases within the training set, as expected since the network is solely trained on those cases. 3) We explored various architectures for training, including fully connected neural networks instead of separate ones. Our conclusions remain consistent; however, it is challenging, if not impossible, to assert that there is not a potentially better NN for improving the data-driven approach.

\newpage
\subsubsection{Influence of the number of training samples}
\begin{figure}[H] 
  \centering
  \includegraphics[width=0.99\linewidth]{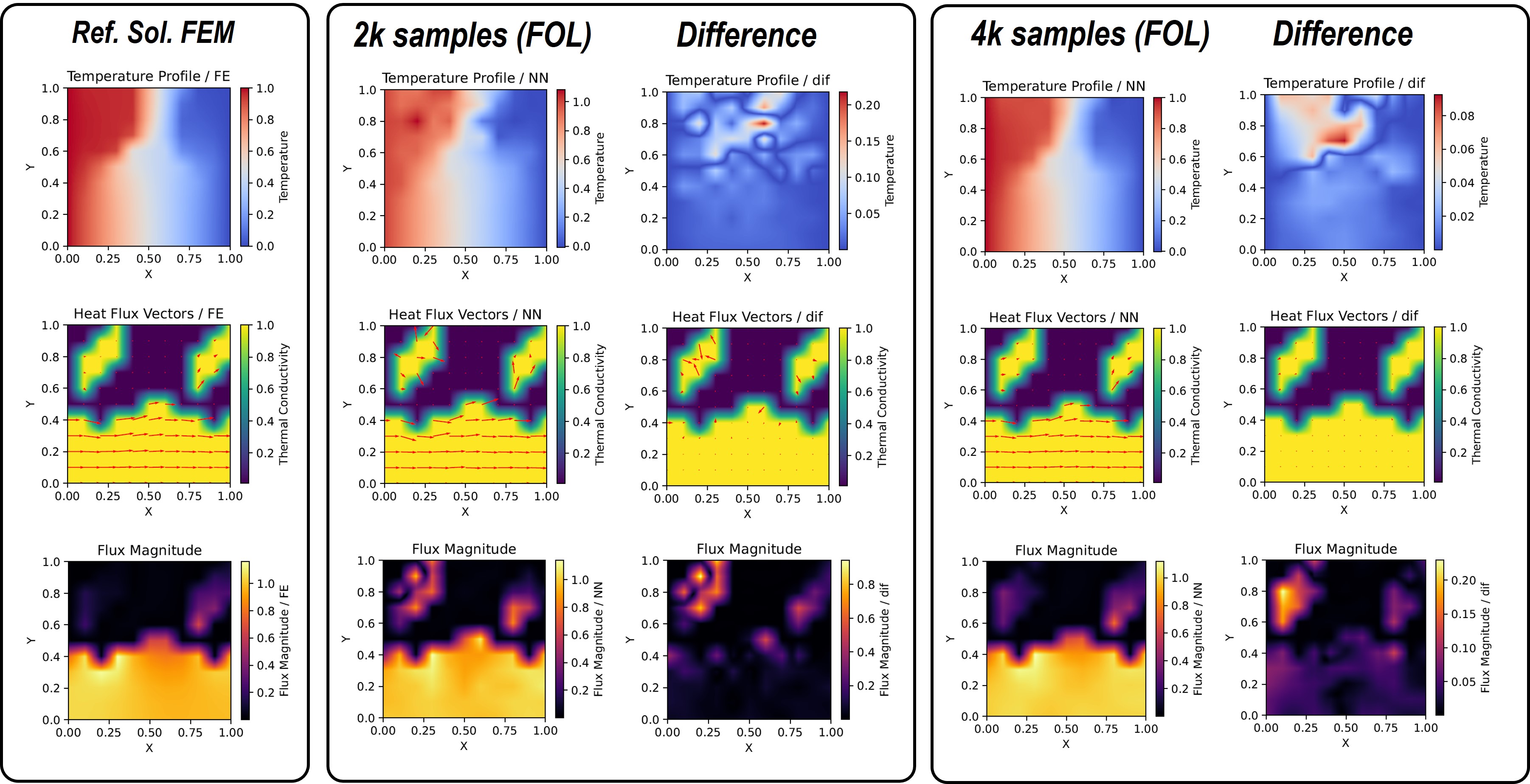}
  \caption{Influence of the initial training fields for the introduced physics-informed deep learning model (FOL).}
  \label{fig:SN_1}
\end{figure}

\begin{figure}[H] 
  \centering
  \includegraphics[width=0.99\linewidth]{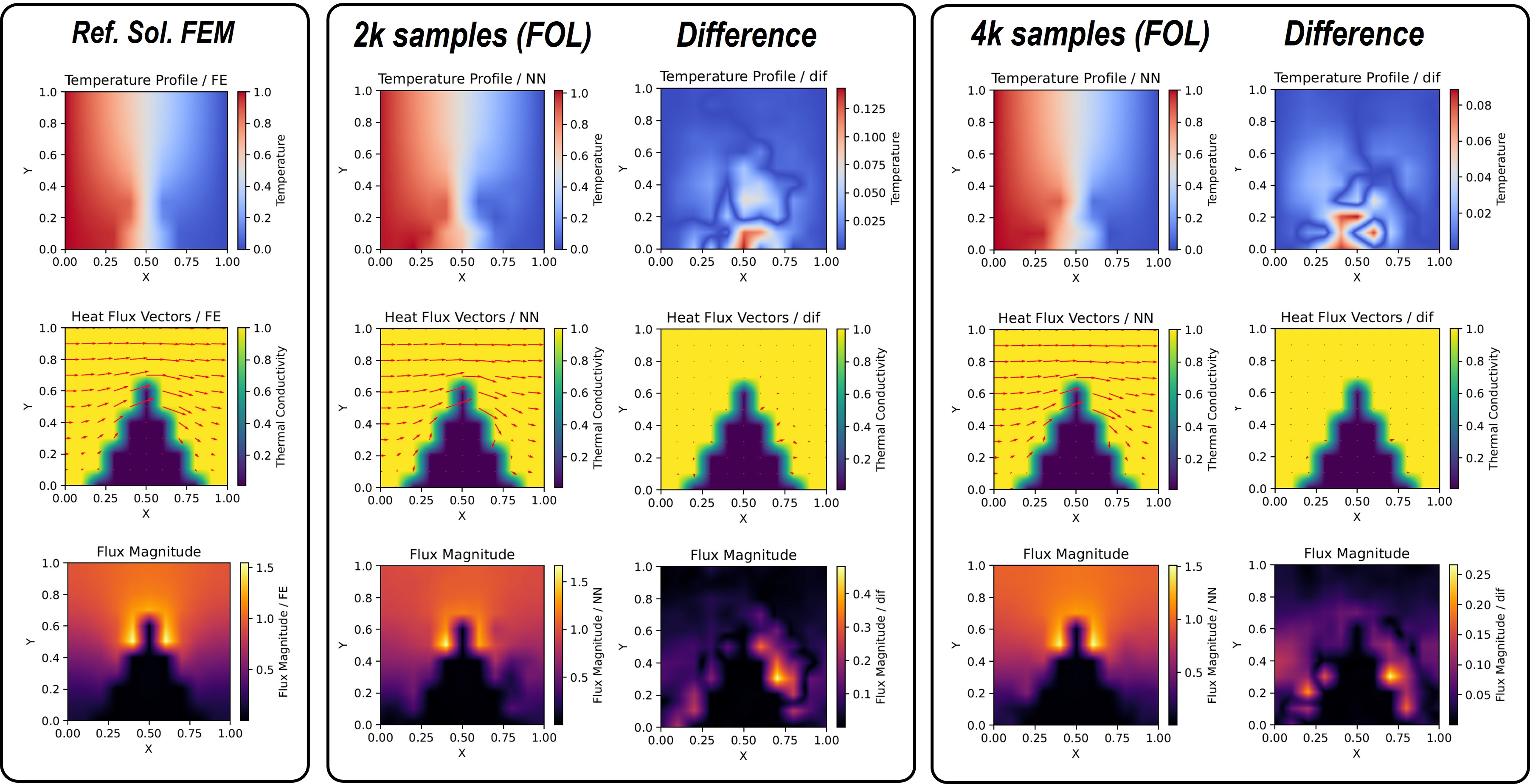}
  \caption{Influence of the initial training fields for the introduced physics-informed deep learning model (FOL).}
  \label{fig:SN_2}
\end{figure}

In Figures \ref{fig:SN_1} and \ref{fig:SN_2}, we analyze the impact of the number of initial sample fields. In the selected 2000 samples, we excluded mainly morphologies with ring shape. For the depicted test cases, we observed a reduction in the maximum difference by a factor of 2.0 and 1.6. This suggests that enhancing the quantity and diversity of initial random fields can improve the quality of results for unseen test cases. Notably, the training cost did not notably increase as we doubled the batch size while increasing the number of samples.

\newpage
\subsubsection{Influence of activation functions}
\begin{figure}[H] 
  \centering
  \includegraphics[width=0.99\linewidth]{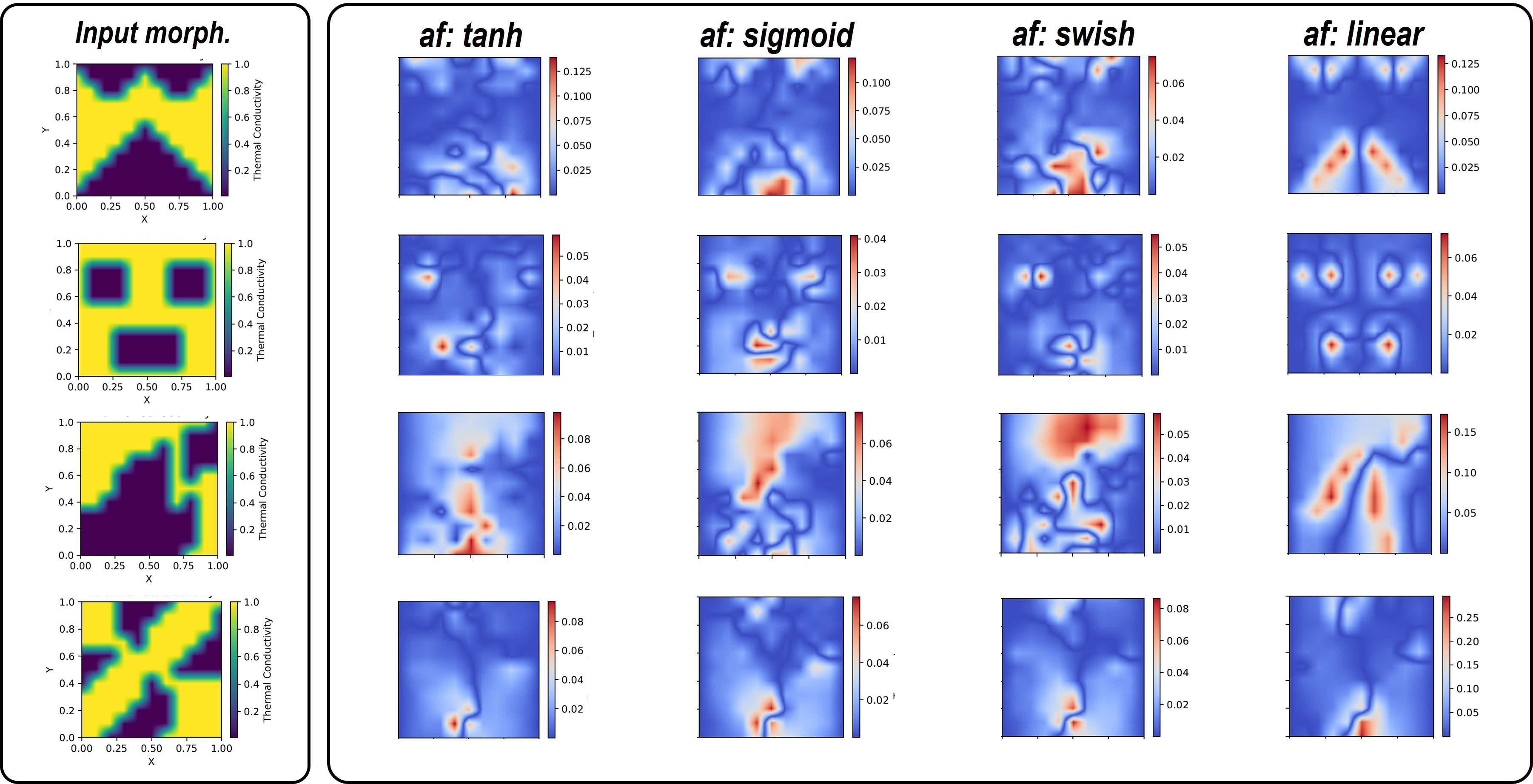}
  \caption{Influence of the type of activation functions on the obtained results.}
  \label{fig:af}
\end{figure}

In Fig.~\ref{fig:af}, we look into the influence of various choices for the activation function on the predictions. To avoid repetition, only the pointwise error for each choice of the activation function is shown in each row and column of Fig.~\ref{fig:af}. Except for the linear case, which is an oversimplified case, the errors using other types of activation functions remain low.

\begin{figure}[H] 
  \centering
  \includegraphics[width=0.8\linewidth]{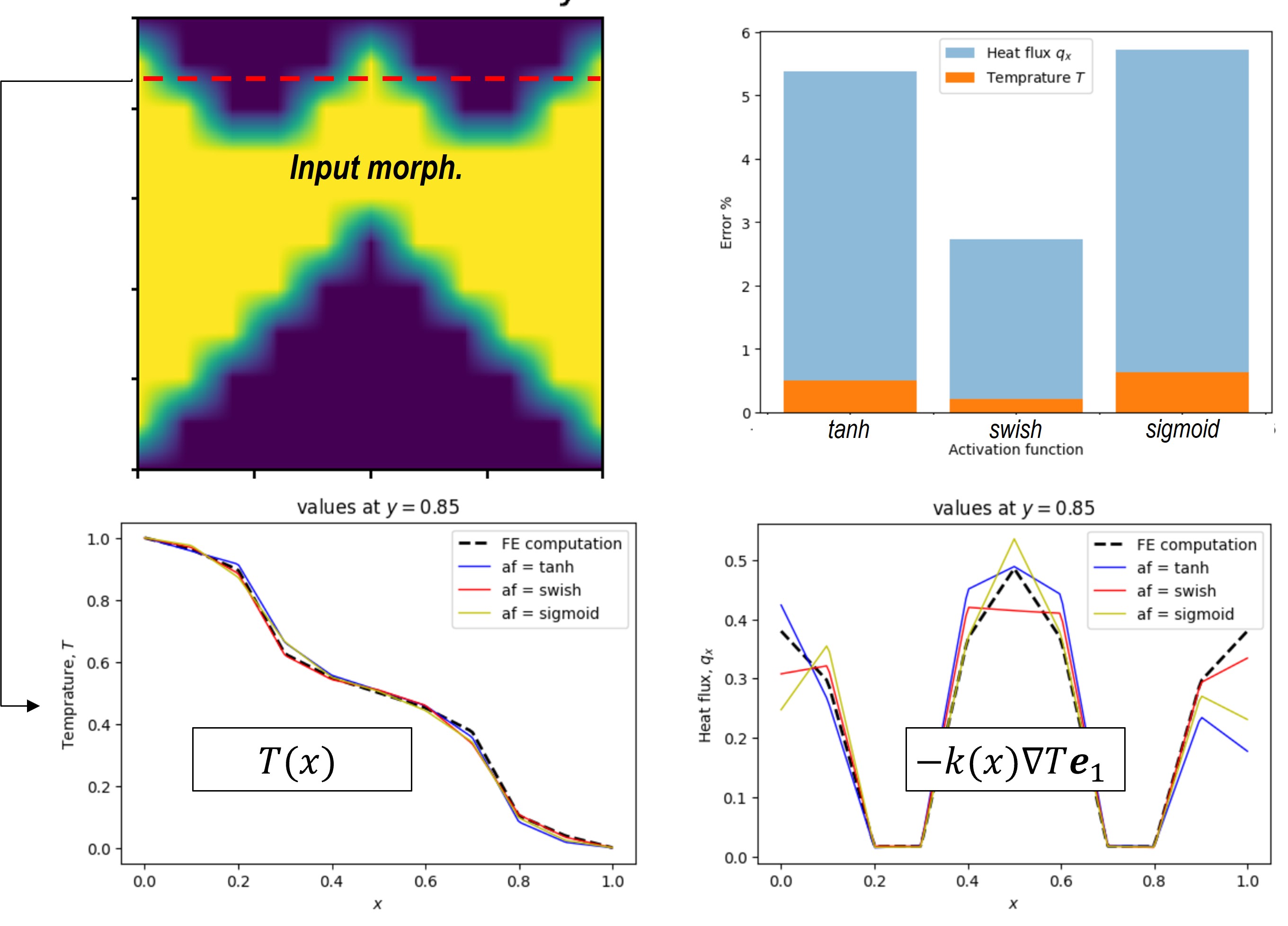}
  \caption{ Evaluation of averaged errors for different choice of activation functions. }
  \label{fig:err}
\end{figure}

In total, the tanh and swish functions showed the best performance. For a better and more quantitative comparison, the cross-section of the predictions and averaged errors are reported in Fig.~\ref{fig:err}. To avoid repetition, the error values in this case are calculated based on $Err = \frac{| \langle T_{NN} \rangle - \langle T_{FE} \rangle|}{| \langle T_{FE} \rangle|}\times 100$, where we have $\langle T_{NN} \rangle = \frac{1}{N^2}\Sigma_i T_{NN}(i)$. This is motivated by the idea of homogenization, where we average the obtained temperature and heat flux to obtain the overall material properties. The lower this homogenized error is, the higher the accuracy will be for the homogenization of properties

Interestingly, the errors for the homogenized temperature are below $1\%$, which also holds for many other test cases not reported here. On the other hand, the average error for the flux in x-direction is higher, which is reasonable since we only predict the temperature profile and any slight fluctuations in the predicted temperature influence its spatial derivatives. The errors for the heat flux using the swish option are still acceptable and remain between $2\%$ to $3\%$. One should note that ideas from mixed formulations, where we predict the heat flux as an output, can also be integrated with the current formulation to enhance the predictive ability of the neural network for higher-order gradients of the solutions (see investigations in \cite{REZAEI2022PINN, Faroughi2023}).

\subsection{Influence of the network architecture}
\begin{figure}[H] 
  \centering
  \includegraphics[width=0.99\linewidth]{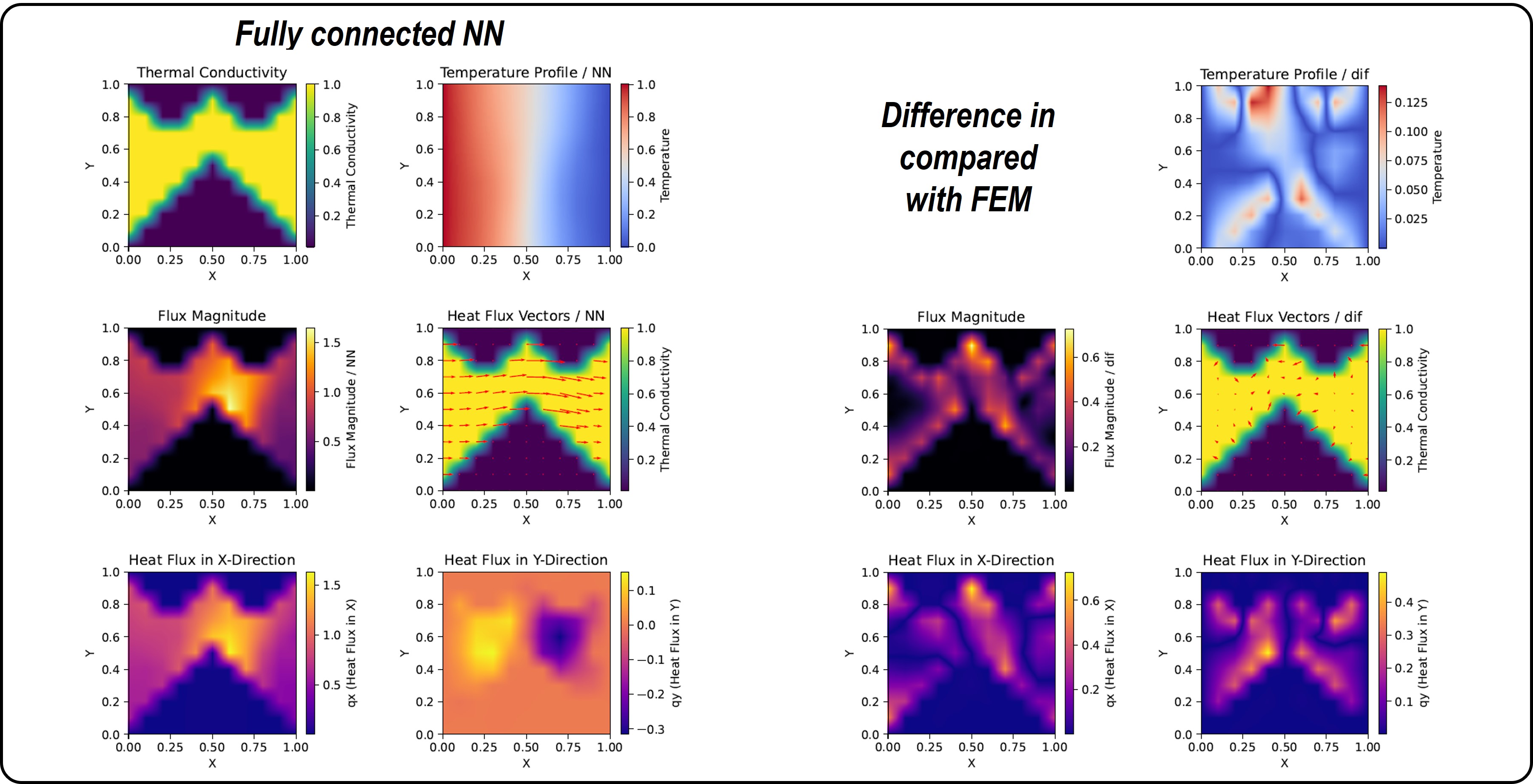}
  \caption{Poor performance of the network by using standard fully connected neural nets instead of separating them. The network parameters such as number of neurons and layers are kept the same as before.}
  \label{fig:NN_arch}
\end{figure}

As discussed in the previous section, the NN architecture is designed to relate all the inputs to one single output. As a result, we are dealing with a parallel series of networks that initially act independently and then are connected through the physics-based loss functions. In this section, we emphasize the importance of this approach by evaluating and training the very same network architecture where we have a fully connected network to map the whole input to the whole output. The number of layers and neurons in each are set to obtain the same numbers from the previous case to ensure a fair comparison. Therefore, we still have 2 layers but with $10 \times 121 = 1210$ neurons in each. The predictions are reported in Fig.~\ref{fig:NN_arch}, where we observe very high errors and rather poor performance from the NN.
It is worth mentioning that the network in this case simply captures the dominant solution, which is a rather linear distribution of temperature from left to right without being able to capture the influence of introduced heterogeneity. Keep in mind that other hyperparameters are also set as before, and even upon changing them for better results, we did not find significant improvements through further parameter studies.

\subsection{Discussions on computational cost and advantages of the FOL}
Thanks to its straightforward formulation, utilizing elementwise domain decomposition from finite elements, and leveraging shape functions, the method facilitates an exceptionally efficient model training process. Table~\ref{tab:network_cost} showcases the required training time per epoch for the thermal problem discussed. As previously mentioned, employing a standard Adam optimizer for roughly 1000 epochs yields satisfactory results. When training the same architecture based on data, the time investment is notably lower, as outlined in Table~\ref{tab:network_cost}. Conversely, once the model is trained, evaluating the neural network becomes remarkably economical for any arbitrary input parameter, taking approximately 1 millisecond on the Apple M2 platform. This evaluation speed is about 20 times faster than the standard FEM. Note that in this paper, we're working with a linear system of equations and a grid of 11 by 11 where the FEM already demonstrates high speed. Consequently, significantly greater acceleration is anticipated for nonlinear problems.

A comprehensive comparison among various available numerical methods is not straightforward due to several reasons. Firstly, some of the newer methods, particularly those based on deep learning, are still under development. Secondly, depending on the application, dimensionality, computational resources, and implementation complexities, certain methods may prove more beneficial than others. Nevertheless, our aim is to summarize and compare the introduced approach in this paper with well-established ones such as the finite element method (FEM) and the standard (vanilla) version of the physics-informed neural network (PINN).
It is worth noting that while PINN methods have undergone further development (e.g., the deep energy method or mixed PINN) and have been enhanced by ideas from transfer learning, these methods are not the primary focus of this comparison. Additionally, among various methods for operator learning (such as physics-based DeepOnet and FNO), further equitable comparisons are needed in future work. Setting these aspects aside, a basic comparison is presented in Table \ref{tab:compare}.


\begin{table}[H]
\centering
\caption{ Comparison of the computational costs per epoch.  }  
\label{tab:network_cost}
\begin{footnotesize}
\begin{tabular}{ l l }
\hline
-   &  normalized averaged run-time     \\
\hline
Training FOL (physics-driven) with 4000 samples, batch size = 100  &  $1.0$  \\
Training FOL (data-driven) with 4000 samples, batch size = 100  &  $0.1$  \\
\hline
Network evaluation after training (data or physics driven)  &   $1.0$  \\ 
Finite element calculation                                  &  $20$  \\ 
\hline
\end{tabular}
\end{footnotesize}
\end{table}
\begin{table}[H]
\fontsize{7}{10}\selectfont
\centering
\caption{ Comparison of different approaches for solving PDEs.}
\label{tab:compare} \begin{adjustbox}{center}
\begin{tabular}{ l l l l }
\hline
---   &  FEM  & PINN &  FOL  \\
\hline
Implementing/Transferring codes  &  Hard (software specific)  & Easy (only network evaluation) & Easy (only network evaluation)  \\
Training time  &   - (doesn't apply)  & Moderate-high (based on AD) & Moderate (based on shape functions) \\
Parametric learning  &  No  & Yes (upon using extra inputs or branch net) & Yes \\
Order of derivations  &  Low (weak form)  & High (unless using DEM or Mixed-PINN) & Low (weak form)  \\
Handling complex geometries  &  Yes  & Yes & Yes \\
Handling strong heterogeneity  &  Yes  & Yes, by using Mixed PINN or domain decomp. & Yes \\
Data requirement  &  Only model param.  & Only model param. & Only model param.  \\ 
&  & no additional solution is required & no additional solution is required \\
\hline
Main limitation  &  Comput. cost + Discretization error  & Training cost + Training range & Training range + Discretization error  \\
Main advantage  &   Accurate solutions for various PDEs  & Fast eval. + No data is required & Fast eval. + No data is required \\
\hline
\end{tabular}
\end{adjustbox}{}
\end{table}

\newpage
\section{Conclusion and outlooks}
This work showcases a simple, data-free, physics-based approach for parametric learning of physical equations across arbitrary-shaped domains. This method relies solely on governing physical constraints, though it can also be augmented with available data. Furthermore, the proposed approach harnesses the advantages of the finite element method without relying on automatic differentiation to compute physical loss functions. Instead, all derivatives are approximated using the concept of shape functions within the finite element framework. This aspect significantly enhances the efficiency of the training process, requiring fewer epochs and less time per epoch. Moreover, it offers flexibility in choosing activation functions without concerns about vanishing gradients. However, it is important to note that potential discretization errors from the finite element method persist here.

We applied our methodology to a steady-state thermal diffusion problem within a two-phase microstructure. The input to the network is the thermal conductivity map, representing the spatial distribution of thermal conductivity, while the network output is the temperature distribution under fixed boundary conditions. Our findings demonstrate that even when considering a relatively high phase contrast, the method accurately predicts the temperature profile. Across all test cases, the maximum pointwise error remained below $10\%$ which is very localized in certain regions such as phase boundaries. The error in the averaged or homogenized temperature across the domain stayed below $1\%$. These levels of error are highly acceptable for numerous engineering applications. However, errors for higher-order gradient terms, such as heat fluxes, tend to increase to higher values, reaching around $5\%$ for averaged errors.

The primary advantage of our current work lies in the ability to train the network in a data-free and parametric manner. Consequently, the network can predict solutions for any arbitrary selection of input parameters, rendering it significantly faster than classical numerical methods that involve repetitively solving algorithms for each parameter set. Additionally, a standard solver does not need to provide data for training. Our findings, particularly for linear systems of equations, exhibited a speed increase ranging from 20 to 50 times, a rate that can potentially escalate further in nonlinear scenarios. Furthermore, upon comparing the results with a data-driven approach, we determined that the physics-driven method offers larger accuracy for unseen input parameters, such as microstructure morphology in this paper.

The current investigations open up an entirely new approach for further research into physics-based operator learning. Expanding this approach to nonlinear problems is not only intriguing but also essential, as it holds the potential for even greater speed improvements. While our focus was primarily on steady-state thermal problems, exploring other types of partial differential equations, such as mechanical equilibrium presents an intriguing avenue for our future scope.
\\ \\ 
\textbf{Data Availability}:
The codes and data associated with this research are available upon request and will be published online following the official publication of the work.
\\ \\
\textbf{Acknowledgements}:
The authors would like to thank the Deutsche Forschungsgemeinschaft (DFG) for the funding support provided to develop the present work in the project Cluster of Excellence “Internet of Production” (project: 390621612). 
\\ \\ 
\textbf{Author Statement}:
S.R.: Conceptualization, Software, Writing - Review \& Editing. A. M.: Software, Review \& Editing. M. K.: Review \& Editing. M. A.: Funding, Review \& Editing.


\bibliography{Ref}

\end{document}